%% file: main.tex
\documentclass[10pt,twocolumn,letterpaper]{article}

\usepackage{cvpr}              


\definecolor{cvprblue}{rgb}{0.21,0.49,0.74}
\usepackage[pagebackref,breaklinks,colorlinks,allcolors=cvprblue]{hyperref}

\usepackage{graphicx}
\usepackage{multirow}
\usepackage{subcaption}
\usepackage[table]{xcolor}
\usepackage[utf8]{inputenc} 
\usepackage[T1]{fontenc}    
\usepackage{hyperref}       
\usepackage{url}            
\usepackage{booktabs}       
\usepackage{amsfonts}       
\usepackage{nicefrac}       
\usepackage{microtype}      
\usepackage{booktabs} 
\usepackage{tabularx}
\usepackage{stfloats}
\usepackage{colortbl}
\usepackage{svg}\usepackage{svg}
\usepackage{flushend}

\usepackage{fontawesome5}
\usepackage{colortbl}
\usepackage{tikz}
\usepackage{ulem}
\usepackage[most]{tcolorbox}
\usepackage{fancyvrb}
\usepackage{fvextra}

\definecolor{cityblue}{RGB}{128, 159, 225}
\definecolor{citypink}{RGB}{227, 108, 194}
\definecolor{w_blue}{RGB}{237, 241, 253}
\usepackage{array,tabularx}
\newcolumntype{C}{>{\centering\arraybackslash}X}
\definecolor{colorbest}{RGB}{252,187,161}
\definecolor{colorsecond}{RGB}{254,224,210}
\definecolor{colorthird}{RGB}{255,245,240}
\definecolor{dataconstruction}{RGB}{109,153,255}
\newcommand{\first}[0]{\cellcolor{colorbest} }
\newcommand{\second}[0]{\cellcolor{colorsecond}}
\newcommand{\third}[0]{\cellcolor{colorthird}}
\DeclareRobustCommand{\legendsquare}[1]{%
  \textcolor{#1}{\rule{2ex}{2ex}}%
}

\newcommand{\ours}{WithAnyone}
\newcommand{\ourdataset}{MultiID-2M}
\newcommand{\ourbenchmark}{MultiID-Bench}

\usepackage{caption}
\title{
\raisebox{-0.015\linewidth}{\includegraphics[width=1.8cm]{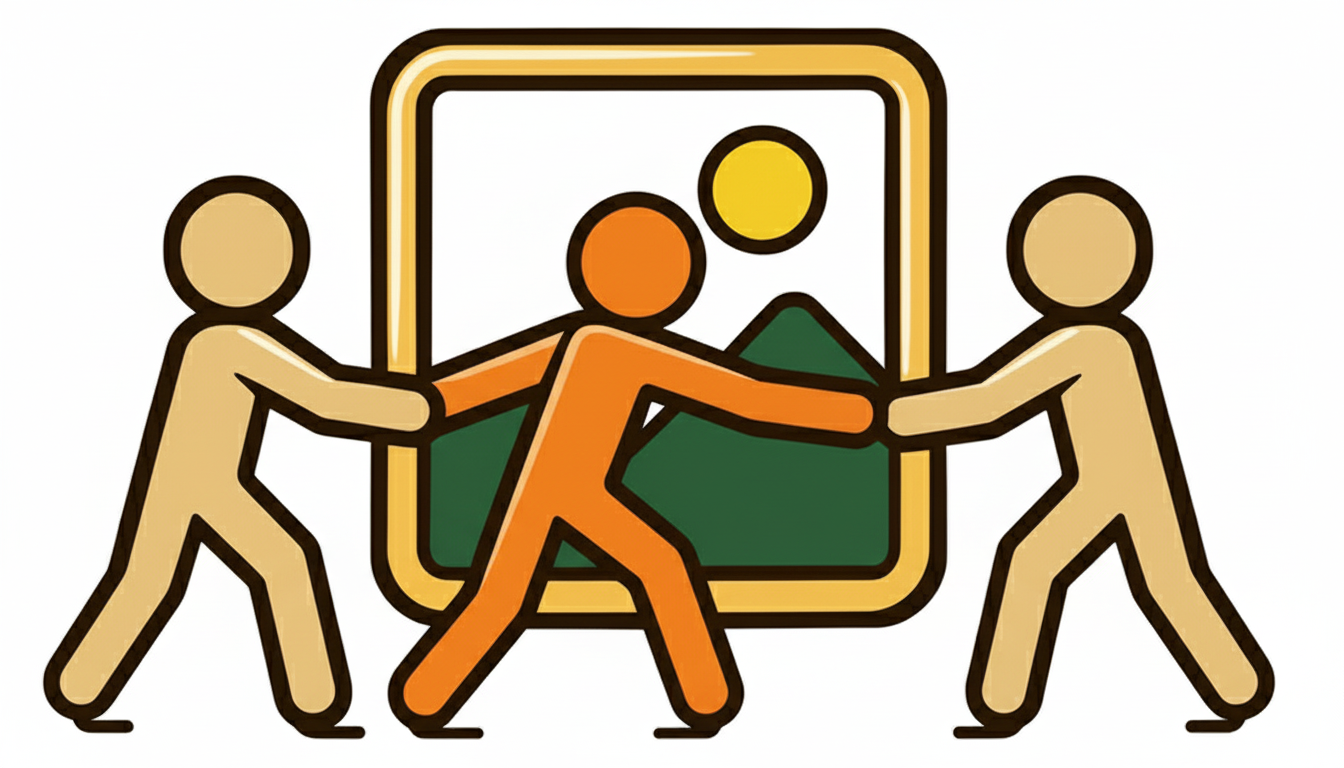}} 
WithAnyone: Towards Controllable and ID Consistent Image Generation}

\author{
    Hengyuan Xu$^{1,2}$
    \quad
    Wei Cheng$^{2,\dag}$
    \quad
    Peng Xing$^{2}$
    \quad
    Yixiao Fang$^{2}$
    \quad
    Shuhan Wu$^{2}$\\
    \quad
    Rui Wang$^{2}$
    \quad
    Xianfang Zeng$^{2}$ 
    \quad
    Daxin Jiang$^{2}$
    \quad
    Gang Yu$^{2,\ddag}$ 
    \quad
    Xingjun Ma$^{1,\ddag}$
    \quad
    Yu-Gang Jiang$^{1}$\\ [0.4em]
    $^{1}$ {Fudan University}
    \quad
    $^{2}$ {StepFun}
    \\[0.4em]
    \textcolor{cityblue}{\normalsize
    \raisebox{-0.2\height}{\includegraphics[height=0.5cm]{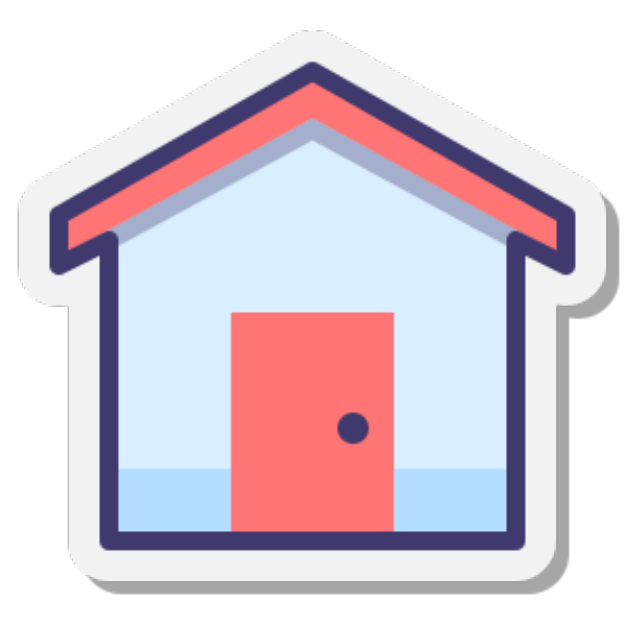}}~{\href{https://doby-xu.github.io/WithAnyone/}{\textbf{Project Page}}}
    \quad
    \raisebox{-0.2\height}{\includegraphics[height=0.5cm]{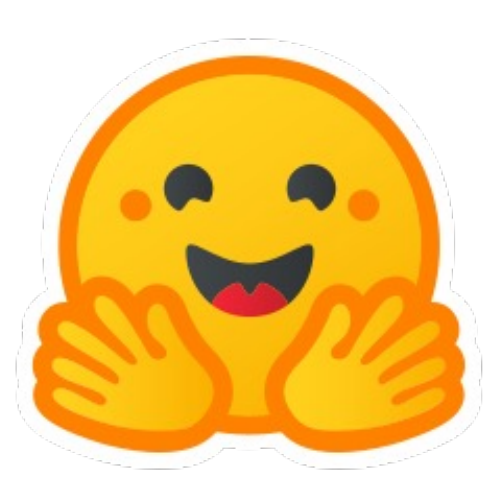}}~{\href{https://huggingface.co/datasets/WithAnyone/MultiID-2M}{\textbf{\ourdataset}}}
    \quad
    \raisebox{-0.2\height}{\includegraphics[height=0.5cm]{files/huggingface_logo.pdf}}~{\href{https://huggingface.co/datasets/WithAnyone/MultiID-Bench}{\textbf{\ourbenchmark}}}
    \quad
    \raisebox{-0.2\height}{\includegraphics[height=0.5cm]{files/huggingface_logo.pdf}}~{\href{https://huggingface.co/WithAnyone/WithAnyone}{\textbf{Models}}}
    \quad
    \raisebox{-0.2\height}{\includegraphics[height=0.5cm]{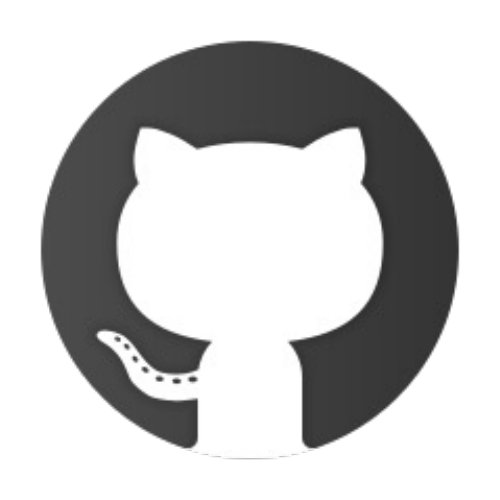}}~{\href{https://github.com/doby-xu/WithAnyone}{\textbf{Code}}}
    } 
}

\begin{document}

\twocolumn[{
  \renewcommand\twocolumn[1][]{#1}
  \maketitle
  \begin{center}
  \vspace{-3ex}
  \includegraphics[width=\textwidth]{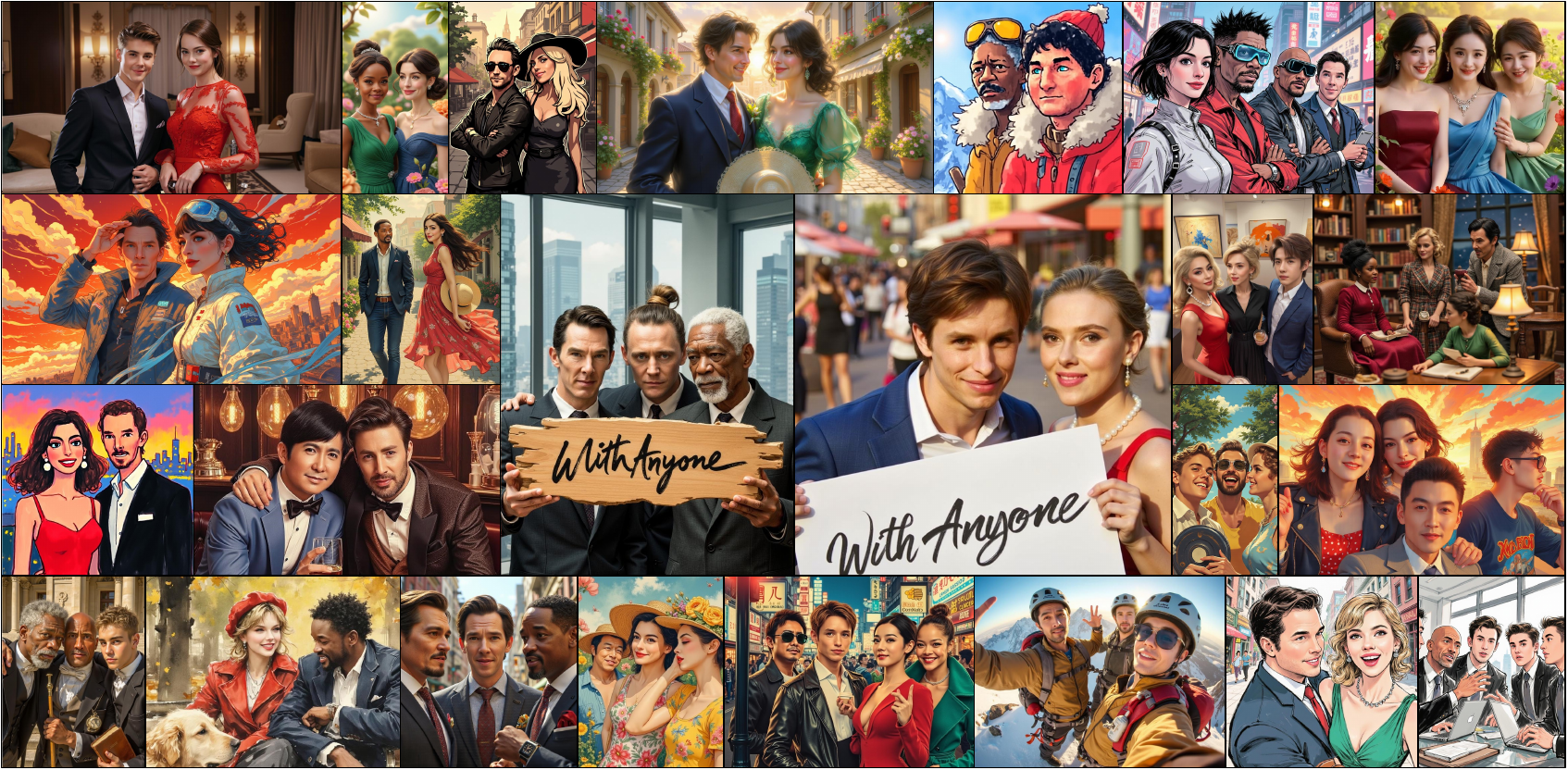}
  \captionof{figure}{\small \textbf{Showcases of \ours.} \ours~is capable of generating high-quality, controllable, and ID-consistent images by leveraging ID-contrastive training on the proposed \textbf{\ourdataset}~dataset.}
  \label{fig:teaser}
  \end{center}
}]

{\let\thefootnote\relax\footnotetext{\noindent$\dag$ Wei Cheng leads this project; \ddag Corresponding authors.}}

\input{sections/0.abstract.tex}

\section{Introduction}\label{sec:intro}

\input{sections/1.introduction.tex}

\section{Related Work}\label{sec:related}
\input{sections/2.related.tex}

\section{\ourdataset : Paired Multi-Person Dataset Construction}\label{sec:dataset}

\input{sections/3.dataset}

\section{\ourbenchmark : Comprehensive ID Customization Evaluation}\label{sec:benchmark}
\input{sections/5.benchmark}

\section{\ours : Controllable and ID-Consistent Generation}\label{sec:method}

\input{sections/4.method}

\section{Experiments}\label{sec:evaluation}
\input{sections/6.evaluation}

\newpage 
\input{sections/7.conclusion}

\clearpage
\newpage
{
\bibliography{main}
\bibliographystyle{plain}
}


\clearpage
\newpage

\input{sections/appendix}

\end{document}

%% file: sections/0.abstract.tex
\begin{abstract}

Identity-consistent generation has become an important focus in text-to-image research, with recent models achieving notable success in producing images aligned with a reference identity. Yet, the scarcity of large-scale paired datasets containing multiple images of the same individual forces most approaches to adopt reconstruction-based training. This reliance often leads to a failure mode we term \textbf{\textit{copy-paste}}, where the model directly replicates the reference face rather than preserving identity across natural variations in pose, expression, or lighting. Such over-similarity undermines controllability and limits the expressive power of generation.
To address these limitations, we (1) construct a large-scale paired dataset \textbf{\ourdataset} tailored for multi-person scenarios, providing diverse references for each identity; (2) introduce a benchmark that quantifies both copy-paste artifacts and the trade-off between identity fidelity and variation; and (3) propose a novel training paradigm with a contrastive identity loss that leverages paired data to balance fidelity with diversity. These contributions culminate in \textbf{\ours}, a diffusion-based model that effectively mitigates copy-paste while preserving high identity similarity.
Extensive qualitative and quantitative experiments demonstrate that \ours~significantly reduces copy-paste artifacts, improves controllability over pose and expression, and maintains strong perceptual quality. User studies further validate that our method achieves high identity fidelity while enabling expressive controllable generation. 

\end{abstract}

%% file: sections/1.introduction.tex
With the rapid progress of generative artificial intelligence, controllable image generation via reference images or image prompting~\citep{ruiz2023dreambooth,hertz2022prompt,zhang2023adding,xiao2024customsketching,hu2025simulating,wu2024fiva} and identity-consistent (ID-consistent) generation~\citep{ye2023ipadapter,guo2024pulid,wang2024instantid,jiang2025infiniteyou,cheng2025umo,zhang2025idpatch,he2024uniportrait} have achieved remarkable advances: modern models can synthesize portraits that closely match the provided individual. Recent efforts~\citep{cheng2025umo,chen2025xverse} push resemblance toward near-perfect reproduction. While pursuing higher similarity seems natural, beyond a certain point, excessive fidelity becomes counterproductive.

In real photographs of the same person, identity similarity varies substantially due to natural changes in pose, expression, makeup, and illumination (Fig.~\ref{fig:observation}). By contrast, many generative models adhere to the reference image far more rigidly than this natural range of variation. Although such over-optimization may seem beneficial, it suppresses legitimate variation, reducing controllability and limiting practical usability. We term this failure mode the \textbf{copy-paste artifact}: rather than synthesizing an identity in a flexible, controllable manner, the model effectively copies the reference image into the output (see Fig.~\ref{fig:observation}). In this work, we formalize this artifact, develop metrics to quantify it, and propose a novel training strategy to mitigate it.

Mitigating copy-paste artifacts is fundamentally constrained by the lack of suitable training data. While numerous large-scale face datasets exist~\citep{liu2015faceattributes,stacchio2020imago,chu2024uniparser,zhang2015beyond,jiang2025referringperson,zhong2018compact,wang2025faceid}, they remain ill-suited for controllable multi-identity generation. Critically, few datasets provide paired references for each identity-multiple images of the same person across diverse expressions, poses, hairstyles, and viewpoints. 
As a result, most prior work resorts to single-person, reconstruction-based training~\citep{guo2024pulid,wang2024instantid}, where the reference and target coincide. This setup inherently promotes copying and exacerbates copy-paste artifacts. Constructing datasets with multiple references per identity, particularly in group photos, and developing methods to effectively exploit such data remain open challenges.

\input{figures/observation}

In this work, we introduce a large-scale open-source Multi-ID dataset, \textbf{\ourdataset}, together with a comprehensive benchmark, \textbf{\ourbenchmark}, designed for intrinsic evaluation of multi-identity image generation. \ourdataset~contains 500k group photos featuring 1–5 recognizable celebrities. For each celebrity, hundreds of individual images are provided as paired references, covering diverse expressions, hairstyles, and viewing angles. In addition, 1.5M unpaired group photos without references are included.
\ourbenchmark~establishes a standardized evaluation protocol for multi-identity generation. Beyond widely adopted metrics such as ID similarity~\citep{schroff2015facenet,deng2019arcface}, it quantifies copy-paste artifacts by measuring distances between generated images, references, and ground truth. Evaluation on 12 state-of-the-art customization models highlights a clear trade-off between ID similarity and copy-paste artifacts (see Fig.~\ref{fig:tradeoff}).

Furthermore, we present \textbf{\ours}, a novel identity customization model built on the FLUX~\citep{flux2024} architecture, as a step toward mitigating copy-paste artifacts. \ours~maintains state-of-the-art identity similarity (with regard to target image) while substantially reducing copy-paste, thereby breaking the long-observed trade-off between fidelity and artifacts. This advance is enabled by a paired-training strategy combined with an ID contrastive loss enhanced with a large negative pool, both made possible by our paired dataset. The labeled identities and their reference images enable the construction of an extended negative pool (images of different identities), which provides stronger discrimination signals during optimization.

In summary, our main contributions are:
\begin{itemize}
\item \textbf{\ourdataset:} A large-scale dataset of 500k group photos containing multiple identifiable celebrities, each with hundreds of reference images capturing diverse variations, along with 1.5M additional unpaired group photos. This resource supports pre-training and evaluation of multi-identity generation models.
\item \textbf{\ourbenchmark:} A comprehensive benchmark with standardized evaluation protocols for identity customization, enabling systematic and intrinsic assessment of multi-identity image generation methods.
\item \textbf{\ours:} A novel ID customization model built on FLUX that achieves state-of-the-art performance, generating high-fidelity multi-identity images while mitigating copy-paste artifacts and enhancing visual quality.
\end{itemize}

%% file: figures/observation.tex
\begin{figure}[htbp]
    \vspace{-3ex}
    \centering
    \includegraphics[width=0.95\linewidth]{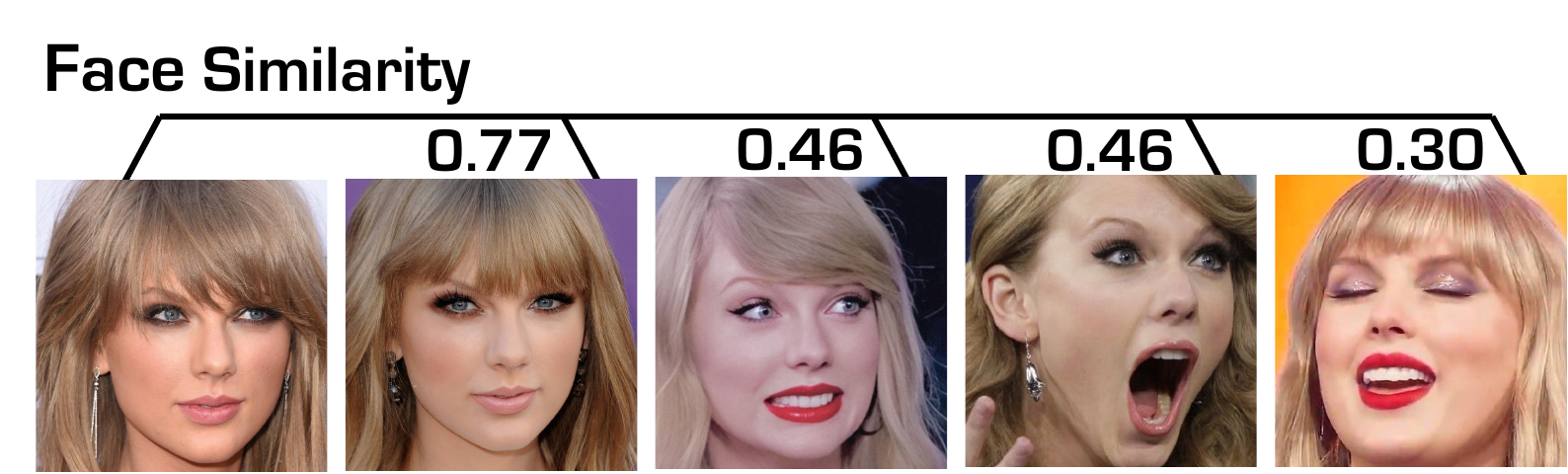}
    \includegraphics[width=0.9\linewidth]{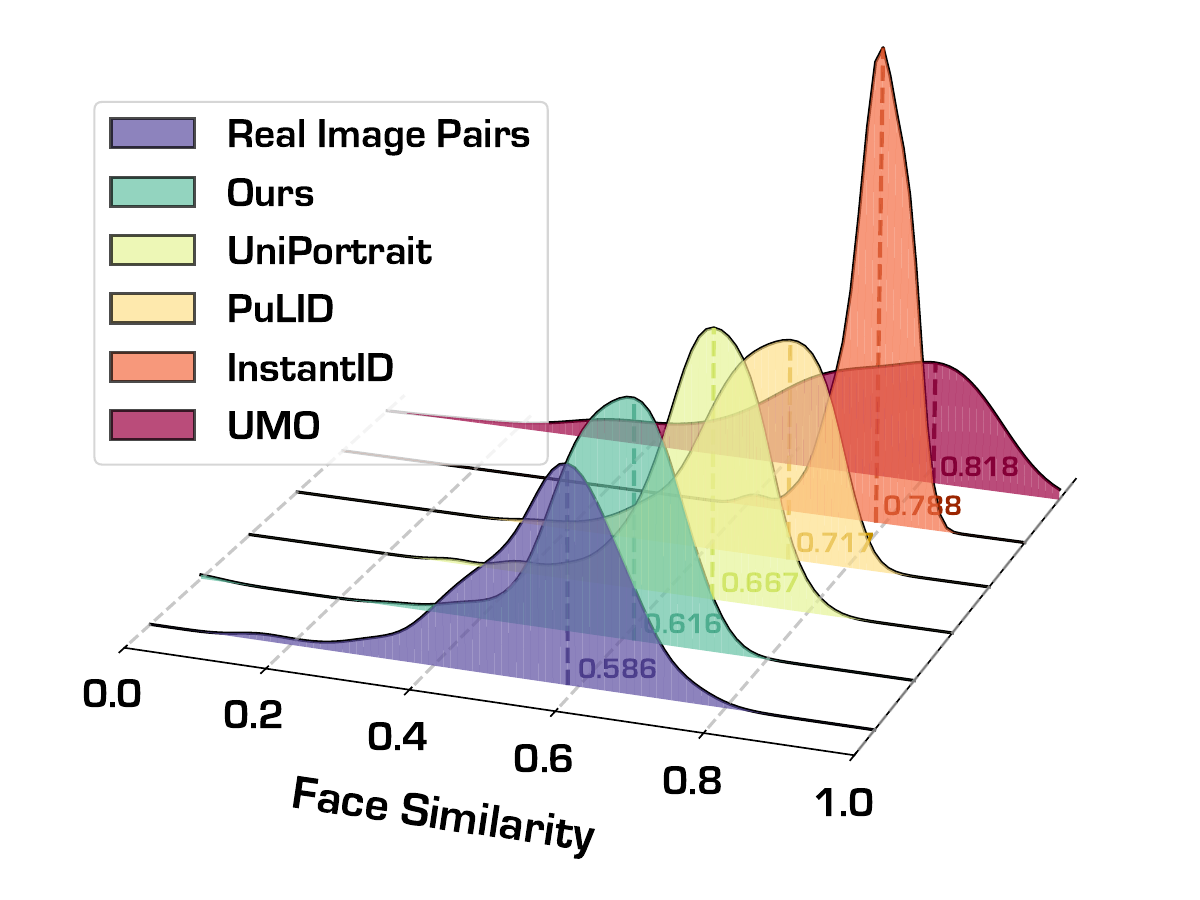}
    \includegraphics[width=1.0\linewidth]{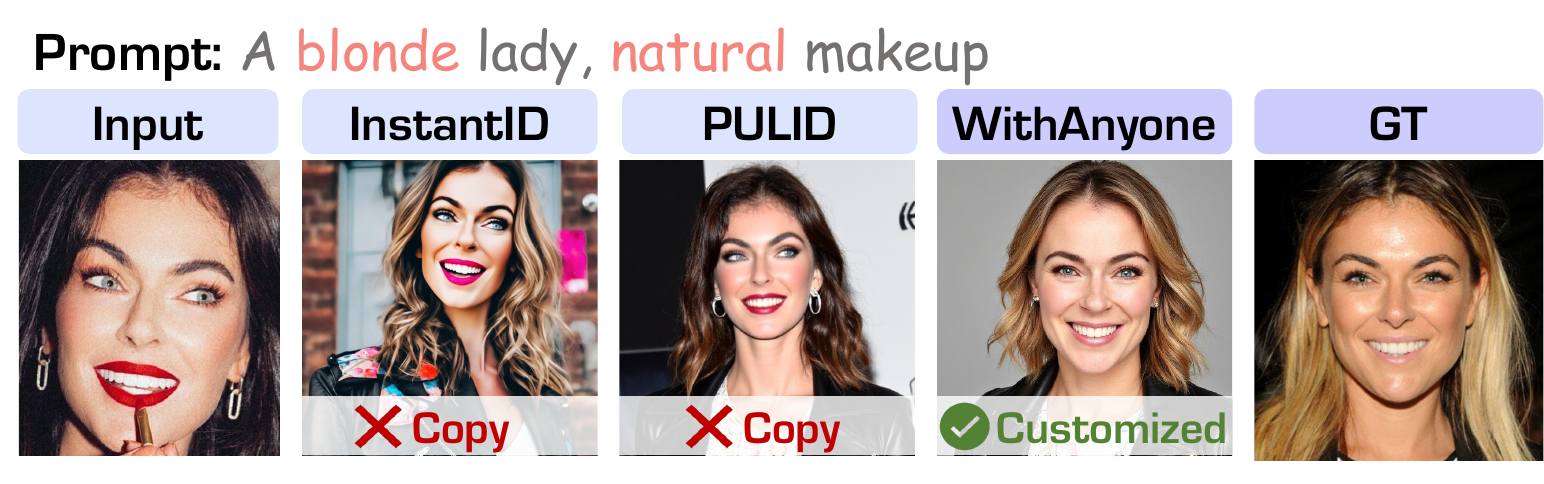}
    \caption{\small \textbf{Our Observation}. Natural variations, such as head pose, expression, and makeup, may cause more face similarity decrease than expected.  Copying reference image limits models' ability to respond to expression and makeup adjustment prompts.}
    \label{fig:observation}
    \vspace{-2ex}
\end{figure}

%% file: sections/2.related.tex
\input{figures/data_ppl}

\paragraph{Single-ID Preservation.}\label{subsec:related_singleid}
The generation of Identity-preserving images is a core topic in customized synthesis~\citep{valevski2023face0,wang2024stableidentity,yan2023facestudio,xiao2025fastcomposer,wu2024infinite,wang2024high,chen2024id,hyung2024magicapture,papantoniou2024arc2face}. Many methods in the UNet/Stable Diffusion era inject learned embeddings (e.g., CLIP or ArcFace) via cross‑attention or adapters~\citep{ho2020ddpm,ronneberger2015u,qian2024omni,ye2023ipadapter,radford2021clip,ren2023pbidr}. With the rise of DiT‑style backbones~\citep{peebles2023scalable,esser2024scaling,flux2024} (e.g., SD3, FLUX), progress in ID preservation like PuLID~\citep{guo2024pulid}, also attracts great attention.

\paragraph{Multi-ID Preservation.}\label{subsec:related_multiid}
Multi‑ID preservation remains relatively underexplored. Some works target spatial control of multiple identities~\citep{kim2024instantfamily,he2024uniportrait,zhang2025idpatch}, while others focus on identity fidelity. Methods such as XVerse~\citep{chen2025xverse} and UMO~\citep{cheng2025umo} use VAE‑derived face embeddings concatenated with model inputs, which can produce pixel‑level copy‑paste artifacts and reduce controllability. DynamicID~\citep{hu2025dynamicid}\footnote{Excluded from our experiments due to unavailability of code and pretrained models.} achieves improved controllability but is constrained by limited task‑specific data and evaluation standards. Other general-purpose customization and editing models~\citep{parmar2025object,mou2025dreamo,patashnik2025nested,wu2025uno,xiao2024omnigen,wu2025omnigen2,wu2025uso,batifol2025flux,wu2025qwenimage} can also synthesize images containing multiple identities, but their ID similarity is often compromised for generality.

\paragraph{ID-Centric Datasets and Benchmarks.}\label{subsec:related_datasets}
Although there are numerous single‑ID datasets \citep{karras2017progressive,wang2025faceid} and multi‑ID collections \citep{chu2024uniparser,jiang2025referringperson}, paired reference images are scarce, so reconstruction remains the dominant training objective for multi‑ID datasets. Representative datasets are listed in Table~\ref{tab:dataset_comparison}. Evaluation protocols are underdeveloped: several works (e.g., PuLID~\citep{guo2024pulid}, UniPortrait~\citep{he2024uniportrait}, and others~\citep{xiao2025fastcomposer,zhang2025idpatch}) construct test sets by sampling identities from CelebA~\citep{liu2015faceattributes}, which undermines reproducibility. Recent efforts benchmark multiple reference generation~\cite{zhuang2025vistorybench,wu2025omnigen2} while focusing on general customization. To address this, we release a curated multi‑ID benchmark with standardized splits and comprehensive metrics to facilitate future research.

%% file: figures/data_ppl.tex
\begin{figure*}[t]
    \centering
    \setlength{\abovecaptionskip}{2pt}
    \includegraphics[width=0.98\linewidth]{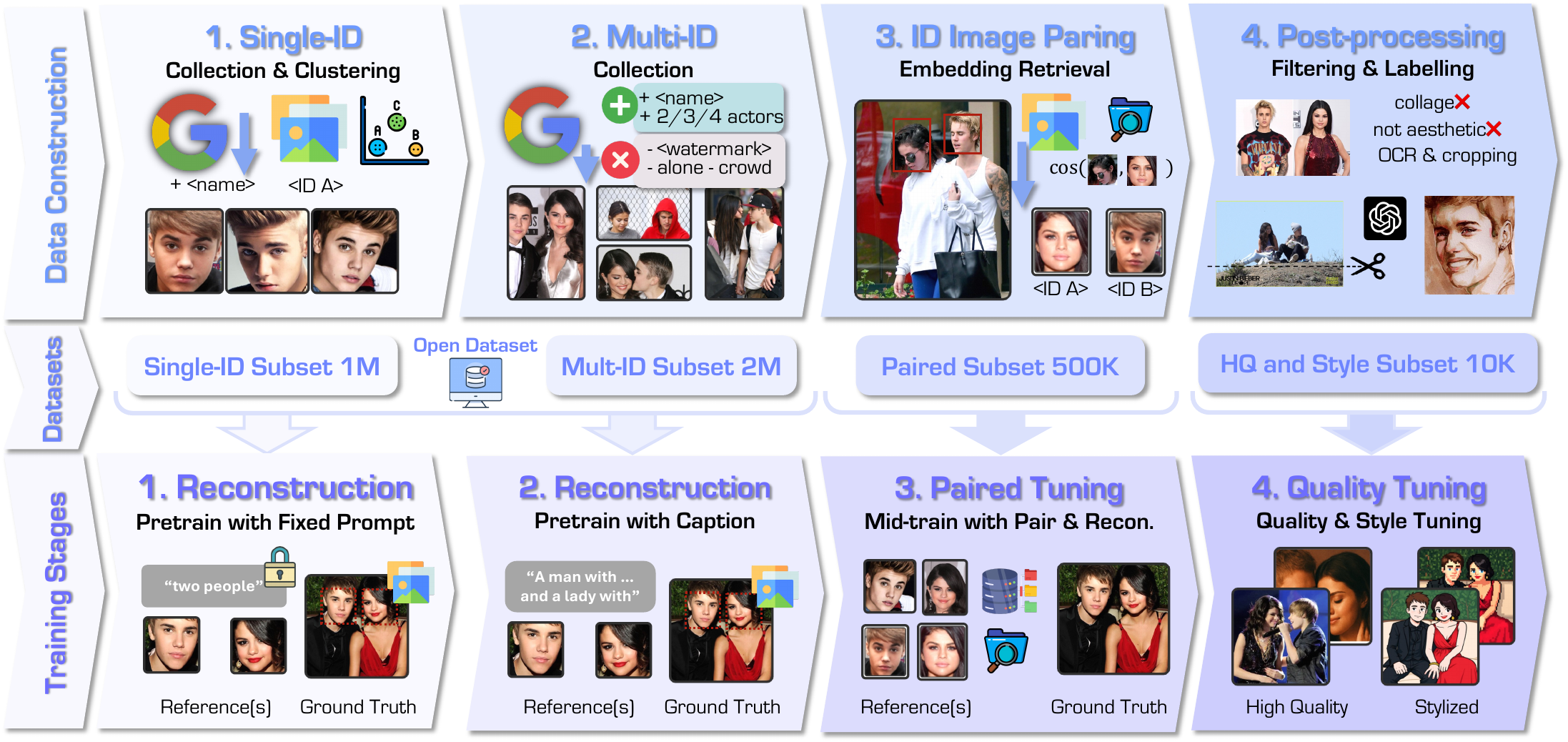}
    \vspace{0.5ex}
    \caption{\small \textbf{Overview of \ours.} It builds on a large-scale dataset, \ourdataset, constructed through a four-step pipeline: (1) collect and cluster single-ID data based on identity similarity; (2) gather multi-ID data via targeted searches using desired identity names with negative keywords for filtering; (3) form image pairs by matching faces between single-ID and multi-ID data; and (4) apply post-processing for quality control and stylization. Training proceeds in four stages: (1) pre-train on single-ID, multi-ID, and open-domain images with fixed prompts; (2) train with image-caption supervision; (3) fine-tune with ID-paired data; and (4) perform quality tuning using a curated high-quality subset.}
    \vspace{-3ex}
    \label{fig:pipeline}
\end{figure*}

%% file: sections/3.dataset.tex
 \ourdataset~is a large-scale multi-person dataset constructed via a four-stage pipeline: (1) collect single-ID images from the web and construct a clean reference bank by clustering ArcFace~\citep{deng2019arcface} embeddings, yielding $\sim $1M reference images across $\sim $3k identities (averaging 400 per identity); (2) retrieve candidate group photos via multi-name and scene-aware queries and detect faces; (3) assign identities by matching ArcFace embeddings to single-ID cluster centers using cosine similarity (threshold 0.4); and (4) perform automated filtering and annotation, including Recognize Anything~\citep{zhang2023recognize}, aesthetic scoring~\citep{aesthetic-predictor-v2-5}, OCR-based watermark/logo removal, and LLM-based caption generation~\citep{Qwen2.5-VL}. The final corpus comprises $\sim $500k identified multi-ID images with matched references from the reference bank, as well as $\sim $1.5M additional unidentified multi-ID images for reconstruction training, covering $\sim $25k unique identities, with diverse nationalities and ethnicities.
Further details of the construction pipeline and dataset statistics are provided in Appendix~\ref{sec:appendix_dataset}.

\input{figures/framework.tex}

%% file: figures/framework.tex
\captionsetup[sub]{labelformat=simple}
\begin{figure*}[t]
    \centering
    \begin{subfigure}[b]{0.51\linewidth}
        \includegraphics[width=\linewidth]{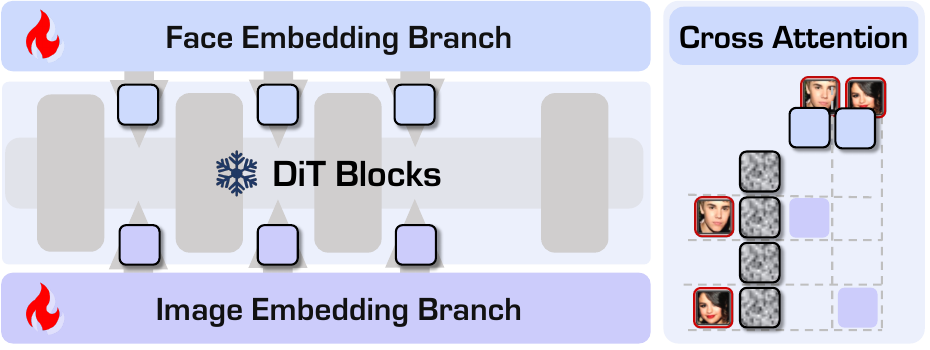}
        \caption{\small \textbf{Model Architecture}}
        \label{fig:tradeoff_single}
    \end{subfigure}
    \begin{subfigure}[b]{0.48\linewidth}
        \includegraphics[width=\linewidth]{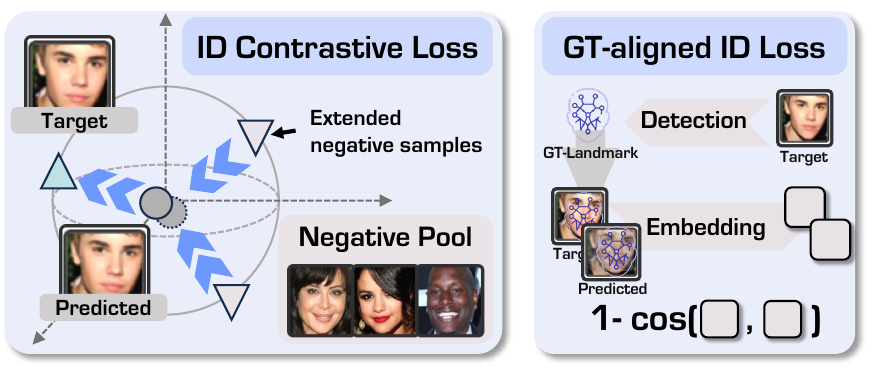}
        \caption{\small \textbf{Training Objectives}}
        \label{fig:tradeoff_multi}
    \end{subfigure}

    \setlength{\abovecaptionskip}{2pt}
    \caption{\small (a) \textbf{Architecture of \ours}: Each reference is encoded by both a face-recognition network and a general image encoder, yielding identity-discriminative signals and complementary mid-level features. Face embeddings are restricted to attend only to image tokens within their corresponding face regions. (b) \textbf{Training Objectives of \ours}: In addition to the diffusion loss, we incorporate an ID contrastive loss and a ground-truth–aligned ID loss, which together provide consistent and accurate identity supervision.}

    \vspace{-1em}
    \label{fig:framework}
\end{figure*}

%% file: sections/5.benchmark.tex
\ourbenchmark~is a unified benchmark for group-photo (multi-ID) generation. It samples rare, long-tail identities with no overlap to training data, yielding 435 test cases. Each case consists of one ground-truth (GT) image containing 1–4 people, the corresponding 1–4 reference images as inputs, and a prompt describing the GT. Detailed statistics are provided in Appendix~\ref{sec:appendix_dataset}.

Evaluation considers both identity fidelity and generation quality. Let $\mathbf{r}, \mathbf{t}, \mathbf{g}$ denote the face embeddings of the reference identity, the target (ground-truth), and the generated image, respectively. We define similarity between two embeddings as $\mathrm{Sim}(\mathbf a,\mathbf b)$, specifically we term the generated image's face similarity with regard to GT as $\mathrm{Sim_{GT}}$, and to reference as $\mathrm{Sim_{Ref}}$,
\begin{equation}\label{eq:identity_fidelity}
\mathrm{Sim}(\mathbf a,\mathbf b) = \frac{\mathbf a^\top \mathbf b}{\|\mathbf a\|\,\|\mathbf b\|}, 
\end{equation}
Specially, we denote $\mathrm{Sim_{Ref}} = \mathrm{Sim}(\mathbf r,\mathbf g)$ and $\mathrm{Sim_{GT}} = \mathrm{Sim}(\mathbf t,\mathbf g)$. 
Prior works~\citep{zhang2025idpatch,he2024uniportrait,guo2024pulid,cheng2025umo} has largely reported only $\mathrm{Sim_{Ref}}$, which inadvertently favors trivial copy-paste: directly replicating the reference appearance maximizes the score, even when the prompt specifies changes in pose, expression, or viewpoint. In contrast, \ourbenchmark~uses $\mathrm{Sim_{GT}}$ the similarity to the ground-truth identity described by the prompt as the primary metric. This design penalizes excessive copying when natural variations (e.g., pose, expression, occlusion) are expected, while rewarding faithful realization of the prompted scene.

We define the angular distance as $\theta_{ab} = \arccos(\mathrm{Sim}(a,b))$ (geodesic distance on the unit sphere). The Copy-Paste metric is given by
\begin{equation}\label{eq:copy-paste}
\mathrm{M_{CP}}(\mathbf g \mid \mathbf t,\mathbf r)
=\frac{\theta_{gt}-\theta_{gr}}{\max(\theta_{tr},\,\varepsilon)} \in [-1,1],
\end{equation}
where $\varepsilon$ is a small constant for numerical stability. The metric thus captures the relative bias of $\mathbf g$ toward the reference $\mathbf r$ versus the ground truth $\mathbf t$, normalized by angular distance of $\mathbf r$ and $\mathbf t$. A score of $1$ means $\mathbf g$ fully coincides with the reference (perfect copy-paste), while $-1$ means full agreement with the ground truth.

We additionally report identity blending, prompt fidelity (CLIP I/T), and aesthetics; formal definitions and further details are provided in Appendix~\ref{sec:appendix_benchmark}.

%% file: sections/4.method.tex
Building on the scale and paired‑reference supervision of the MultiID‑2M, we devise training strategies and tailored objectives that transcend reconstruction to enable robust, identity‑conditioned synthesis. This rich, identity‑labeled supervision not only substantially improves identity fidelity but also suppresses trivial copy–paste artifacts and affords finer control over multi‑identity composition. Motivated by these advantages, we introduce WithAnyone - a unified architecture and training recipe designed for controllable, high‑fidelity multi‑ID generation. Architectural schematics and implementation details are provided in Fig.~\ref{fig:framework} and Appendix~\ref{sec:model_framework}.

\subsection{Training Objectives}
\label{sec:training_loss}

\textbf{Diffusion Loss.} We adopt the mini-batch empirical flow-matching loss. For each batch, we sample a data latent $x_1 \sim p_{\text{data}}$, Gaussian noise $x_0 \sim \mathcal{N}(0,I)$, and a timestep $t \sim \mathcal{U}(0,1)$. We then form the interpolated latent $x_t = (1-t)x_0 + t x_1$ and regress the target velocity $(x_1 - x_0)$:
\begin{equation}\label{eq:diffusion_loss}
\mathcal{L}_{\mathrm{diff}} = \big\| v_\theta(x_t^{(i)}, t^{(i)}, c^{(i)}) - (x_1^{(i)} - x_0^{(i)}) \big\|_2^2,
\end{equation}
where $c^{(i)}$ denotes the conditioning signal.

\input{tables/quantative}

\textbf{Ground-truth-Aligned ID Loss.} Since ArcFace embedding requires landmark detection and alignment, directly extracting landmarks from $I_{\text{gen}}$ is unreliable because generated images are obtained through noisy diffusion or one-step denoising. Prior methods compromise: PortraitBooth~\citep{peng2024portraitbooth} applies the loss only at low noise levels ($t<0.25$), discarding supervision at higher noise, while PuLID~\citep{guo2024pulid} fully denoises generated results at significant computational cost. In contrast, we align the generated image using GT landmarks, thereby avoiding noisy landmark extraction. We minimize the cosine distance between GT-aligned ArcFace embeddings of the generated and ground-truth (GT) faces:
\begin{equation}\label{eq:id_loss}
\mathcal{L}_{\text{ID}} = 1 - \cos(\mathbf g, \mathbf t)
\end{equation}
where $\mathbf g$ and $\mathbf t$ are ArcFace embeddings of the generated and GT images. This design (1) enables applying the ID loss across all noise levels, (2) incurs negligible overhead throughout training, and (3) implicitly supervises generated landmarks. Ablation studies (Sec.~\ref{sec:ablation-study}) demonstrate more accurate identity measurement and substantially improved identity preservation.

Denoting the face recognition model as $f(\cdot, \cdot)$ (Arcface~\citep{deng2019arcface}, in our case), and the coupled detection model as $g(\cdot)$ (RetinaFace~\citep{Deng2020retinaface}), the generated image as $\mathbf G$, and the ground-truth image as $\mathbf T$, a embedding extraction should be performed as follows:
\begin{equation}
    \mathbf t = f(g(\mathbf T), \mathbf T),
\end{equation}
where $g(\mathbf T)$ are the detected landmarks, and $f(\cdot, \cdot)$ extracts the aligned face embedding. Instead of using $g(\mathbf G)$ as landmarks for $\mathbf G$, our GT-aligned ID loss is computed as:
\begin{equation}
    \mathcal{L}_{\text{id}} = 1 - \cos(f(g(\mathbf T), \mathbf G),  f(g(\mathbf T), \mathbf T)).
\end{equation}

\textbf{ID Contrastive Loss With Extended Negatives.} To further strengthen identity preservation, we introduce an ID contrastive loss that explicitly pulls the generated image closer to its reference images in the face embedding space while pushing it away from other identities. The loss follows the InfoNCE~\citep{oord2018representation} formulation:
\begin{equation}
\mathcal{L}_{\text{CL}} = -\log \frac{\exp(\cos(\mathbf g, \mathbf r)/\tau)}{\sum_{j=1}^{M} \exp(\cos(\mathbf g, \mathbf n_j))/\tau)},
\end{equation}
where $\mathbf r$ is the embedding of a reference image of the same identity as the generated image, $\mathbf n_j$ are embeddings of $M$ negatives from different identities, and $\tau$ is a temperature hyperparameter. This formulation relies on ID-labeled datasets, which make it possible to draw thousands of negatives per sample from the reference bank, thereby greatly enriching the diversity of negative examples.

The overall training objective is a weighted sum of the above losses:
\begin{equation}
    \mathcal{L} = \mathcal{L}_{\mathrm{diff}} + \lambda_{\text{ID}} \mathcal{L}_{\text{ID}} + \lambda_{\text{CL}} \mathcal{L}_{\text{CL}},
\end{equation}
where $\lambda_{\text{ID}}$ and $\lambda_{\text{CL}}$ are hyper-parameters controlling the contributions of the ID loss and contrastive loss, respectively. Both are set to $0.1$ across all training phases described below.


\input{figures/trade_off.tex}
\input{tables/quantative_multi}

\subsection{Training pipeline}
\label{sec:training_pipeline}

Copy–paste artifacts largely arise from reconstruction-only training, which encourages models to replicate the reference image rather than learn robust identity-conditioned generation. Leveraging our paired dataset, we employ a four-phase training pipeline that gradually transitions the objective from reconstruction toward controllable, identity-preserving synthesis.

\textbf{Phase 1: Reconstruction pre-training with fixed prompt.} We begin with reconstruction pre-training to initialize the backbone, as this task is simpler than full identity-conditioned generation and can exploit large-scale unlabeled data. For the first few thousand steps, the caption is fixed to a constant dummy prompt (e.g., “two people”), ensuring the model prioritizes learning the identity-conditioning pathway rather than drifting toward text-conditioned styling. The full \ourdataset~is used in this phase, which typically lasts for $20\text{k}$ steps, at which point the model achieves satisfactory identity similarity. To further enhance data diversity, CelebA-HQ~\citep{karras2017progressive}, FFHQ~\citep{karras2019style}, and a subset of FaceID-6M~\citep{wang2025faceid} are also incorporated.

\textbf{Phase 2: Reconstruction pre-training with full captions.} This phase aligns identity learning with text-conditioned generation and lasts for an additional $40\text{k}$ steps, during which the model reaches peak identity similarity.

\textbf{Phase 3: Paired tuning.} To suppress trivial copy–paste behavior, we replace $50\%$ of the training samples with paired instances drawn from the $500\text{k}$ labeled images in \ourdataset. For each paired sample, instead of using the same image as both input and target, we randomly select one reference image from the identity’s reference set and another distinct image of the same identity as the target. This perturbation breaks the shortcut of direct duplication and compels the model to rely on high-level identity embeddings rather than low-level copying.

\textbf{Phase 4: Quality tuning.} Finally, we fine-tune on a curated high-quality subset augmented with generated stylized variants to (i) enhance perceptual fidelity and (ii) improve style robustness and transferability. This phase refines texture, lighting, and stylistic adaptability while preserving the strong identity consistency established in earlier phases.

%% file: tables/quantative.tex
\begin{table*}[b]
    \centering
    \caption{\small \textbf{Quantitative comparison on the single-person subset of  \ourbenchmark~and OmniContext}. \legendsquare{colorbest}, \legendsquare{colorsecond}, and \legendsquare{colorthird} indicate the first-, second-, and third-best performance, respectively. For Copy-Paste ranking, only cases with $\mathrm{Sim(GT)} > 0.40$ are considered.}
    \begin{subtable}[t]{0.60\textwidth}
        \centering
        \caption{\small \textbf{\ourbenchmark}}
        \scalebox{0.70}{
        \begin{tabular}{l|ccc|ccc}
        \toprule
        \multirow{2}[2]{*}{\textbf{Method}} & \multicolumn{3}{c|}{\textbf{Identity Metrics}} & \multicolumn{3}{c}{\textbf{Generation Quality}} \\
        \cmidrule(r){2-4} \cmidrule(l){5-7}
         & Sim(GT) $\uparrow$ & Sim(Ref) $\uparrow$ &  CP $\downarrow$ & CLIP-I $\uparrow$ & CLIP-T $\uparrow$ & Aes $\uparrow$ \\
        \midrule
        DreamO & 0.454 & 0.694 & 0.303 & 0.793 & 0.322 & 4.877 \\
        OmniGen & 0.398 & 0.602 & 0.248 & 0.780 & 0.317 & 5.069 \\
        OmniGen2 & 0.365 & 0.475 & 0.142 & 0.787 & \first{0.331} & 4.991 \\
        FLUX.1 Kontext & 0.324 & 0.408 & 0.099 & 0.755 & 0.327 & \third{5.319} \\
        Qwen-Image-Edit & 0.324 & 0.409 & 0.093 & 0.776 & 0.316 & 5.056 \\
        GPT-4o Native & 0.425 & 0.579 & \second{0.178} & \second{0.794} & 0.311 & \second{5.344} \\
        UNO & 0.304 & 0.428 & 0.141 & 0.765 & 0.314 & 4.923 \\
        USO & 0.401 & 0.635 & 0.286 & 0.790 & \second{0.329} & 5.077 \\
        UMO & \third{0.458} & \second{0.732} & 0.359 & 0.783 & 0.305 & 4.850 \\
        \midrule

        UniPortrait & 0.447 & 0.677 & 0.265 & \third{0.793} & 0.319 & 5.018 \\
        ID-Patch & 0.426 & 0.633 & \third{0.231} & 0.792 & 0.312 & 4.900 \\
        InfU & 0.439 & 0.630 & 0.233 & 0.772 & \third{0.328} & \first{5.359} \\
        PuLID  & 0.452 & \third{0.705} & 0.315 & 0.779 & 0.305 & 4.839 \\
        InstantID  & \first{0.464} & \first{0.734} & 0.337 & 0.764 & 0.295 & 5.255 \\
        Ours & \second{0.460} & 0.578 & \first{0.144} & \first{0.798} & 0.313 & 4.783 \\
        \midrule
        GT & 1.000 & 0.521 & -0.999 & N/A & N/A & N/A \\
        Ref & 0.521 & 1.000 & 0.999 & N/A & N/A & N/A \\
        \bottomrule
    \end{tabular}
        }
    \end{subtable}\hfill
    \begin{subtable}[t]{0.40\textwidth}
    \centering
        \caption{\small \textbf{OmniContext Single Character Subset} }
    \scalebox{0.80}{
        \begin{tabular}{l|cc|c}
        \toprule
        \multirow{2}[2]{*}{\textbf{Method}} & \multicolumn{2}{c|}{\textbf{Quality Metrics}} & \textbf{Overall} \\
        \cmidrule(r){2-3} \cmidrule(l){4-4}
         & PF $\uparrow$ & SC $\uparrow$ & Overall $\uparrow$ \\
        \midrule
        DreamO & 8.13 & 7.09 & 7.02 \\
        OmniGen & 7.50 & 5.52 & 5.47 \\
        OmniGen2 & 8.64 & 8.50 & 8.34 \\
        FLUX.1 Kontext & 7.72  & 8.60 & 7.94 \\
        Qwen-Image-Edit & 7.66 & 8.16 & 7.51 \\
        GPT-4o Native & 7.98 & 9.06 & 8.12 \\
        UNO & 7.22 & 7.72 & 7.04\\
        USO & 6.96 & 7.88 & 6.70 \\
        UMO         & 6.56 & 7.92 & 6.79 \\
        \midrule
        
        UniPortrait & 6.62 & \third{6.00} & \third{5.55} \\
        ID-Patch    & N/A & N/A & N/A \\
        InfU       & \first{7.69} & 4.62 & 4.70 \\
        PuLID    & \third{6.62} & \second{6.83} & \second{5.78} \\
        InstantID  & 4.89 & 5.49 & 4.35 \\
        Ours        & \second{7.43} & \first{7.04} & \first{6.52} \\
        \bottomrule
    \end{tabular}
    
        }
    \end{subtable}

    \label{tab:quantitative_comparison_single}
\end{table*}

%% file: figures/trade_off.tex
\captionsetup[sub]{labelformat=simple}
\begin{figure*}[t]
    \centering
    \begin{subfigure}[b]{0.48\linewidth}
        \includegraphics[width=\linewidth]{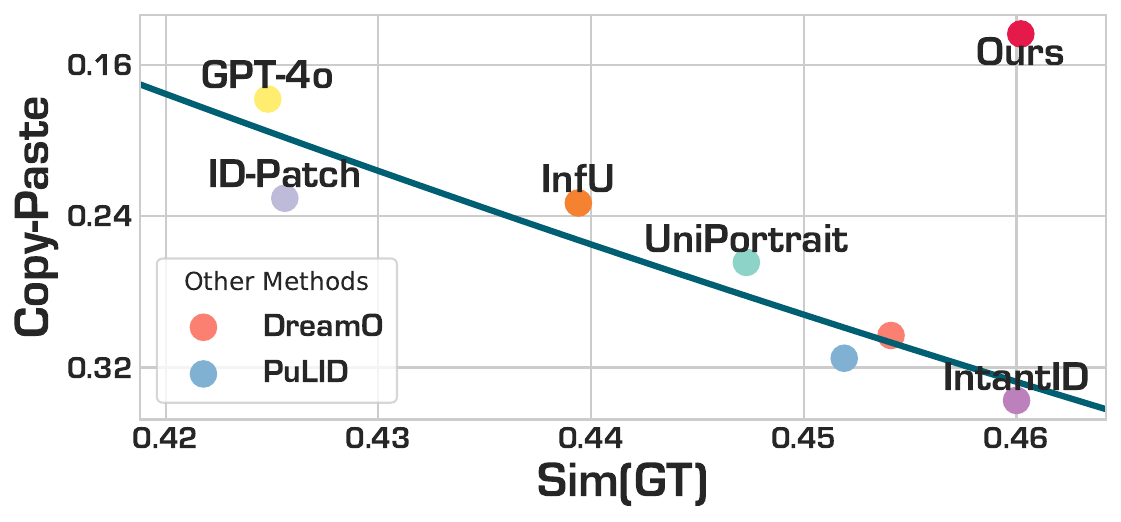}
        \caption{Single-ID subset}
        \label{fig:tradeoff_single}
    \end{subfigure}
    \hspace{2ex}
    \begin{subfigure}[b]{0.48\linewidth}
        \includegraphics[width=\linewidth]{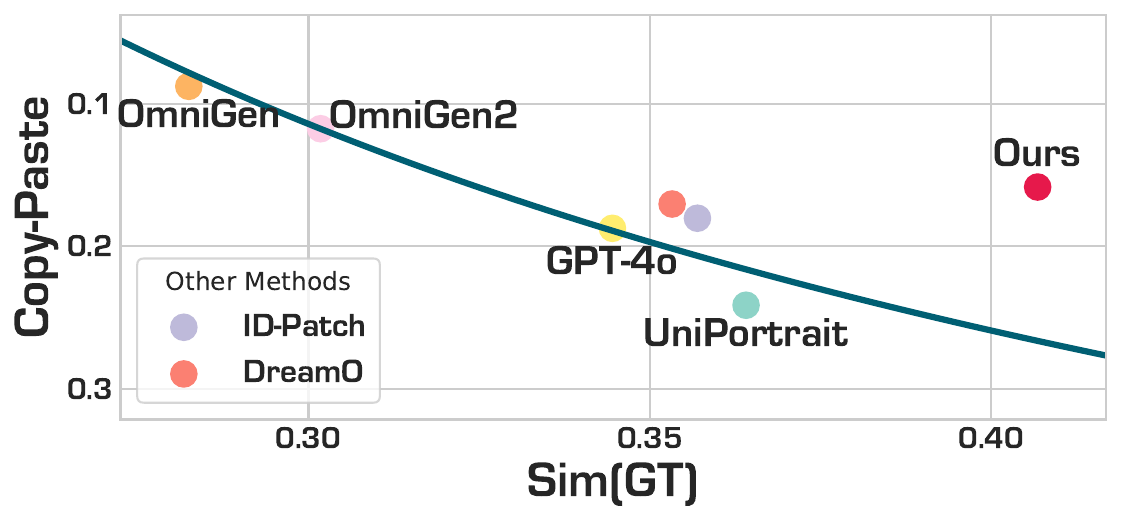}
        \caption{Multi-ID subset}
        \label{fig:tradeoff_multi}
    \end{subfigure}

    \setlength{\abovecaptionskip}{2pt}
    \caption{\textbf{Trade-off between Face Similarity and Copy-paste.} Except for \ours, the other models fall roughly on a fitted curve, illustrating a clear trade-off between face similarity and copy-paste. Upper-right corner is desired.}

    \label{fig:tradeoff}
\end{figure*}

%% file: tables/quantative_multi.tex
\begin{table*}[b]
    \centering
    \caption{\small \textbf{Quantitative comparison on the multi-person subset of \ourbenchmark}. \legendsquare{colorbest}, \legendsquare{colorsecond}, and  \legendsquare{colorthird} indicate the first-, second-, and third-best performance, respectively. For Copy-Paste ranking, only cases with $\mathrm{Sim(GT)} > 0.35$ are considered. GPT exhibits prior knowledge of identities from TV series in subsets with more than two IDs, leading to abnormally high similarity scores.}
    \begin{subtable}[t]{0.48\textwidth}
        \centering
        \caption{\small \textbf{2-people Subset}}
    \scalebox{0.60}{
    \begin{tabular}{l|cccc|ccc}
        \toprule
        \multirow{2}[2]{*}{\textbf{Method}} & \multicolumn{4}{c|}{\textbf{Identity Metrics}} & \multicolumn{3}{c}{\textbf{Generation Quality}} \\
        \cmidrule(r){2-5} \cmidrule(l){6-8}
         & Sim(GT) $\uparrow$ & Sim(Ref) $\uparrow$ &  CP $\downarrow$ &  Bld $\downarrow$ & CLIP-I $\uparrow$ & CLIP-T $\uparrow$ & Aes $\uparrow$ \\
        \midrule
        DreamO & \third{0.359} & 0.514 & \second{0.179} & 0.105 & 0.763 & 0.319 & 4.764 \\
        OmniGen & 0.345 & \third{0.529} & 0.209 & 0.110 & 0.750 & 0.326 & \third{5.152} \\
        OmniGen2 & 0.283 & 0.353 & 0.081 & 0.112 & 0.763 & \first{0.334} & 4.547 \\
        GPT & 0.332 & 0.400 & 0.061 & 0.092 & \first{0.774} & \second{0.328} & \first{5.676} \\
        UNO & 0.223 & 0.274 & 0.043 & \third{0.082} & 0.735 & 0.325 & 4.805 \\
        \midrule
        UMO& 0.328 & 0.491 & 0.176 & 0.111 & 0.743 & 0.316 & 4.772 \\
        
        UniPortrait & \second{0.367} & \first{0.601} & 0.254 & \first{0.075} & 0.750 & 0.323 & \second{5.187} \\
        ID-Patch & 0.350 & 0.517 & \third{0.183} & 0.085 & \third{0.767} & \third{0.326} & 4.671 \\
        Ours & \first{0.405} & \second{0.551} & \first{0.161} & \second{0.079} & \second{0.770} & 0.321 & 4.883 \\
        \bottomrule
    \end{tabular}
    }
    \end{subtable}
    \begin{subtable}[t]{0.48\textwidth}
        \centering
        \caption{\small \textbf{3-and-4-people Subset}}
    \scalebox{0.60}{
    \begin{tabular}{l|cccc|ccc}
        \toprule
        \multirow{2}[2]{*}{\textbf{Method}} & \multicolumn{4}{c|}{\textbf{Identity Metrics}} & \multicolumn{3}{c}{\textbf{Generation Quality}} \\
        \cmidrule(r){2-5} \cmidrule(l){6-8}
       & Sim(GT) $\uparrow$ & Sim(Ref) $\uparrow$ &  CP $\downarrow$ &  Bld $\downarrow$ & CLIP-I $\uparrow$ & CLIP-T $\uparrow$ & Aes $\uparrow$ \\
        \midrule
        DreamO      & 0.311 & 0.427 & 0.116 & 0.081 & 0.709 & 0.317 & 4.695 \\
        OmniGen     & \third{0.345} & \third{0.529} & 0.209 & 0.110 & 0.750 & 0.326 & \second{5.152} \\
        OmniGen2    & 0.288 & 0.374 & 0.099 & 0.071 & 0.734 & \second{0.329} & 4.664 \\
        GPT        & \textit{0.445} & \textit{0.484} & \textit{0.048} & \textit{0.044} & \first{0.815} & \third{0.320} & \first{5.647} \\
        UNO         & 0.228 & 0.276 & 0.046 & 0.065 & 0.717 & 0.319 & 4.880 \\
        \midrule
        UMO         & 0.318 & 0.465 & 0.180 & 0.070 & 0.717 & 0.309 & 4.946 \\
        
        UniPortrait & 0.343 & 0.517 & 0.178 & \second{0.048} & 0.708 & 0.323 & \third{5.090} \\
        ID-Patch    & \second{0.379} & \second{0.543} & \second{0.195} & \third{0.059} & \second{0.781} & \first{0.329} & 4.547 \\
        Ours        & \first{0.414} & \first{0.561} & \first{0.171} & \first{0.045} & \third{0.771} & 0.325 & 4.955 \\
        \bottomrule
    \end{tabular}
    }
    \end{subtable}
    \vspace{-1em}
    \label{tab:quantitative_comparison_multi}
\end{table*}

%% file: sections/6.evaluation.tex
\input{figures/showcase}

In this section, we present a comprehensive evaluation of baselines and our \ours~model on the proposed \ourbenchmark.

\textbf{Baselines.}
We evaluate two categories of baseline methods: general customization models and face customization methods. The general customization models include OmniGen~\citep{xiao2024omnigen}, OmniGen2~\citep{wu2025omnigen2}, Qwen-Image-Edit~\citep{wu2025qwenimage}, FLUX.1 Kontext~\citep{batifol2025flux}, UNO~\citep{wu2025uno}, USO~\citep{wu2025uso}, UMO~\citep{cheng2025umo}, and native GPT-4o-Image~\citep{openai2025gpt4o}. The face customization methods include UniPortrait~\citep{he2024uniportrait}, ID-Patch~\citep{zhang2025idpatch},  PuLID~\citep{guo2024pulid} (referring to its FLUX~\citep{flux2024} implementation throughout this paper), and InstantID~\citep{wang2024instantid}. All models were evaluated on the single-person subset of the benchmark, while only those supporting multi-ID generation were additionally tested on the multi-person subset.
Further implementation details are provided in Appendix~\ref{sec:implementation-details}.

\subsection{Quantitative Evaluation}\label{sec:quantitative-evaluation}

The quantitative results are reported in Tables~\ref{tab:quantitative_comparison_single} and \ref{tab:quantitative_comparison_multi}. We observe a clear trade-off between face similarity and copy-paste artifacts. As shown in Fig.~\ref{fig:tradeoff}, most methods align closely with a regression curve, where higher face similarity generally coincides with stronger copy-paste. This indicates that many existing models boost measured similarity by directly replicating reference facial features rather than synthesizing the identity. In contrast, \ours~deviates substantially from this curve, achieving the highest face similarity with regard to GT while maintaining a markedly lower copy-paste score.

\ours~also achieves the highest score among ID-specific reference models on the OmniContext~\citep{wu2025omnigen2} benchmark. However, VLMs~\citep{Qwen2.5-VL,openai2025gpt4o} exhibit limited ability to distinguish individual identities and instead emphasize non-identity attributes such as pose, expression, or background. Despite that general customization and editing models often outperform face customization models on OmniContext, \ours~still has best performance among face customization models.

\subsection{Qualitative Comparison}

To complement the quantitative results, Fig.~\ref{fig:showcases} presents qualitative comparisons between our method, state-of-the-art general customization/editing models, and face customization generation models.

It shows that identity consistency remains a significant weakness of general customization or editing models, consistent with our quantitative findings. Many VAE-based approaches where references are encoded through a VAE, such as FLUX.1 Kontext and DreamO tend to produce faces that either exhibit copy-paste artifacts or deviate markedly from the target identity. A likely reason is that VAE embeddings emphasize low-level features, leaving high-level semantic understanding to the diffusion backbone, which may not have been pre-trained for this task.
ID-specific reference models also struggle with copy-paste artifacts. For example, they fail to make the subject smile when the reference image is neutral and often cannot adjust head pose or even eye gaze. In contrast, \ours~generates flexible, controllable faces while faithfully preserving identity.

\subsection{Ablation and User Studies}
\label{sec:ablation-study}
\input{tables/abl_cp_data.tex}
\input{figures/gt_aligned_lmk}

To better understand the contribution of each component in \ours, we conduct ablation studies on the training strategy, the GT-aligned ID loss, the InfoNCE-based ID loss, and our dataset. Due to space constraints, we report the key results here, with additional analyses provided in Appendix~\ref{sec:appendix_ablation}.

As shown in Table~\ref{tab:abl_cp_data}, the paired-data fine-tuning phase reduces copy-paste artifacts without diminishing similarity to the ground truth, while training on FFHQ performs significantly worse than on our curated dataset. Fig.~\ref{fig:gt_aligned_lmk} further demonstrates that the GT-aligned ID loss lowers denoising error at low noise levels and yields higher-variance, more informative gradients at high noise, thereby strengthening identity learning. By ablating extended negatives, leaving only $63$ negative samples from the batch (originally extended to $4096$), the effectiveness of ID contrastive loss is greatly reduced. More ablation results can be found in Appendix~\ref{sec:appendix_ablation}.

We conduct a user study to evaluate perceptual quality and identity preservation. Ten participants were recruited and asked to rank 230 groups of generated images according to four criteria: identity similarity, presence of copy-paste artifacts, prompt adherence, and aesthetics. The results, shown in Fig.~\ref{fig:user_study}, indicate that our method consistently achieves the highest average ranking across all dimensions, demonstrating both stronger identity preservation and superior visual quality. Moreover, the copy-paste metric exhibits a moderate positive correlation with human judgments, suggesting that it captures perceptually meaningful artifacts. Further details of the study design, ranking protocol, and statistical analysis are provided in Appendix~\ref{sec:user_study}.

\input{figures/user_study}

%% file: figures/showcase.tex

\begin{figure*}[t]
    \centering
    \includegraphics[width=0.98\textwidth]{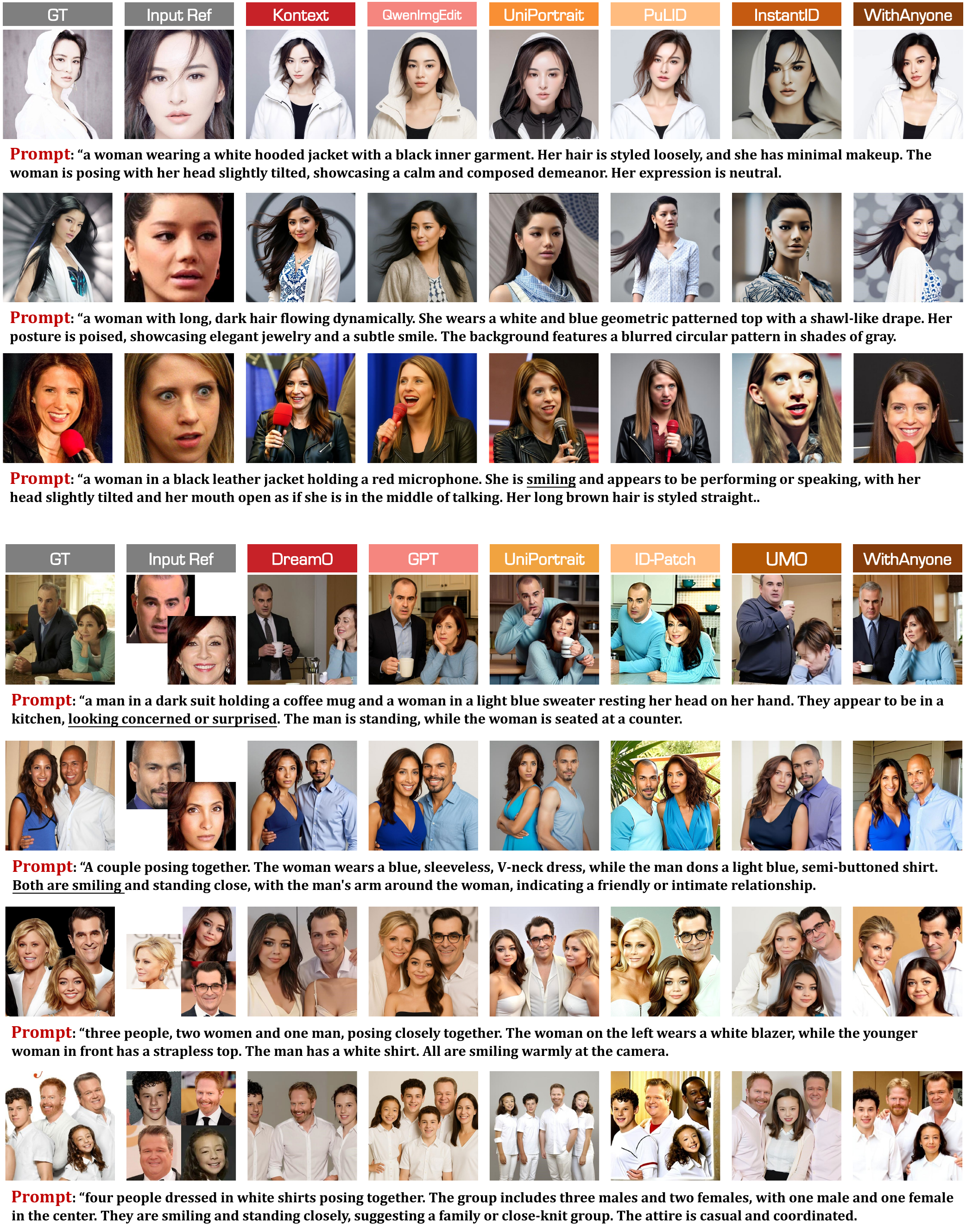}
    \caption{\small \textbf{Qualitative Results of Different Generation Methods.} The text prompt is extracted from the ground-truth image shown on the leftmost side.}
    \vspace{-1em}
    \label{fig:showcases}
\end{figure*}

%% file: tables/abl_cp_data.tex
\begin{table}[htbp]
    \centering
    \caption{\textbf{Ablation Study}. \legendsquare{colorbest} \legendsquare{colorsecond} \legendsquare{colorthird} indicate the first, second, third performance respectively. We ablate paired data training (without stage 2, w/o s2), GT-Aligned landmark ID loss (Self-aligned, S.A.), extended negative samples in InfoNCE (w/o neg). And model trained on FFHQ is also compared.}
    \scalebox{0.60}{
    \begin{tabular}{l|l|ccc|ccc}
            \toprule
            \multirow{2}{*}{} & \multirow{2}[2]{*}{\textbf{Ablation}} & \multicolumn{3}{c|}{\textbf{Identity Metrics}} & \multicolumn{3}{c}{\textbf{Generation Quality}} \\
            \cmidrule(r){3-5} \cmidrule(l){6-8}
             &  & Sim(G) $\uparrow$ & Sim(R) $\uparrow$ & CP $\downarrow$ & CLIP-I $\uparrow$ & CLIP-T $\uparrow$ & Aes $\uparrow$ \\
             \midrule
             
            Phases
                & w/o Phase 3 & \first{0.406} & \first{0.625} & 0.239 & \third{0.755} & 0.307 & \third{4.955} \\
            \midrule
            \multirow{2}{*}{Loss} 
                & w/o GT-Align   & \third{0.385} & \third{0.549} & \third{0.175} & \second{0.763} & \third{0.317} & 4.754 \\
                & w/o Ext. Neg. & 0.368 & 0.455 & \first{0.074} & 0.740 & 0.304 & \second{4.984} \\
            \midrule
            \multirow{1}{*}{Data} 
                & FFHQ only & 0.224 & 0.246 & 0.027 & 0.658 & \first{0.330} & \first{5.039} \\
            \midrule
            \multirow{1}{*}{Ours} 
                & Full Setting & \second{0.405} & \second{0.551} & \second{0.161} & \first{0.770} & \second{0.321} & 4.883 \\
            
            \bottomrule
        \end{tabular}
    }
    \label{tab:abl_cp_data}
\end{table}

%% file: figures/gt_aligned_lmk.tex
\begin{figure}[h]
    \centering
    \includegraphics[width=0.44\textwidth]{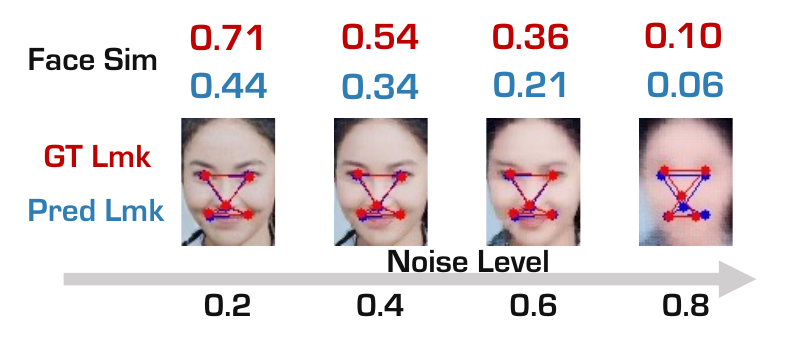}
    \includegraphics[width=0.38\textwidth]{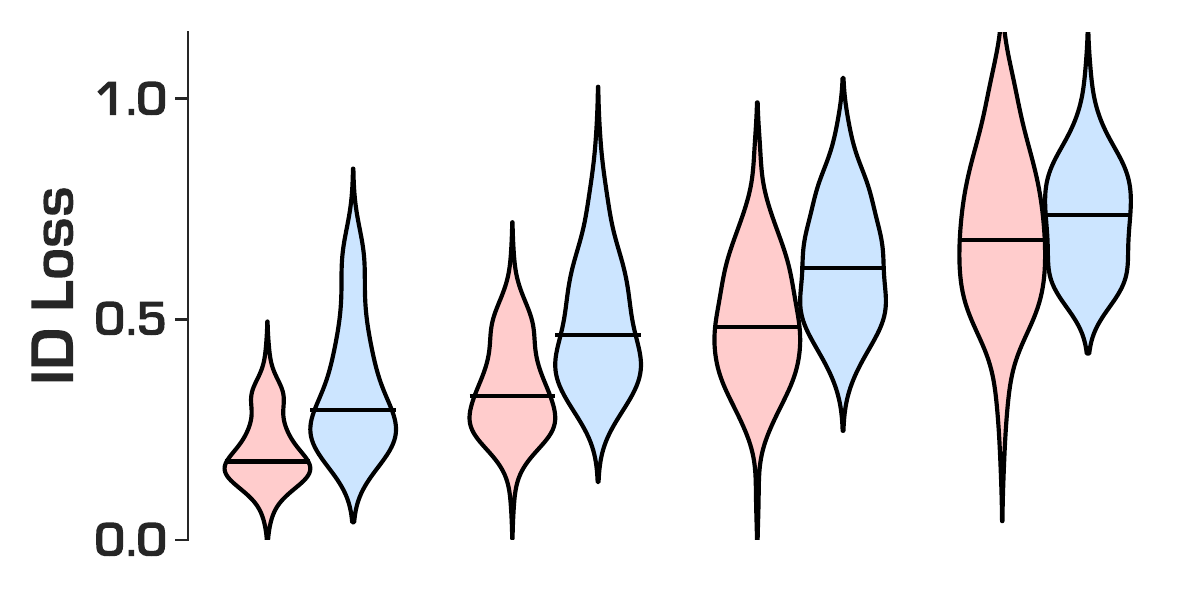}
    \caption{\textbf{Comparison of GT-aligned and Prediction-aligned landmarks.} }
    \label{fig:gt_aligned_lmk}
\end{figure}

%% file: figures/user_study.tex
\begin{figure}[htbp]
    \centering
        \centering
        \includegraphics[width=\linewidth]{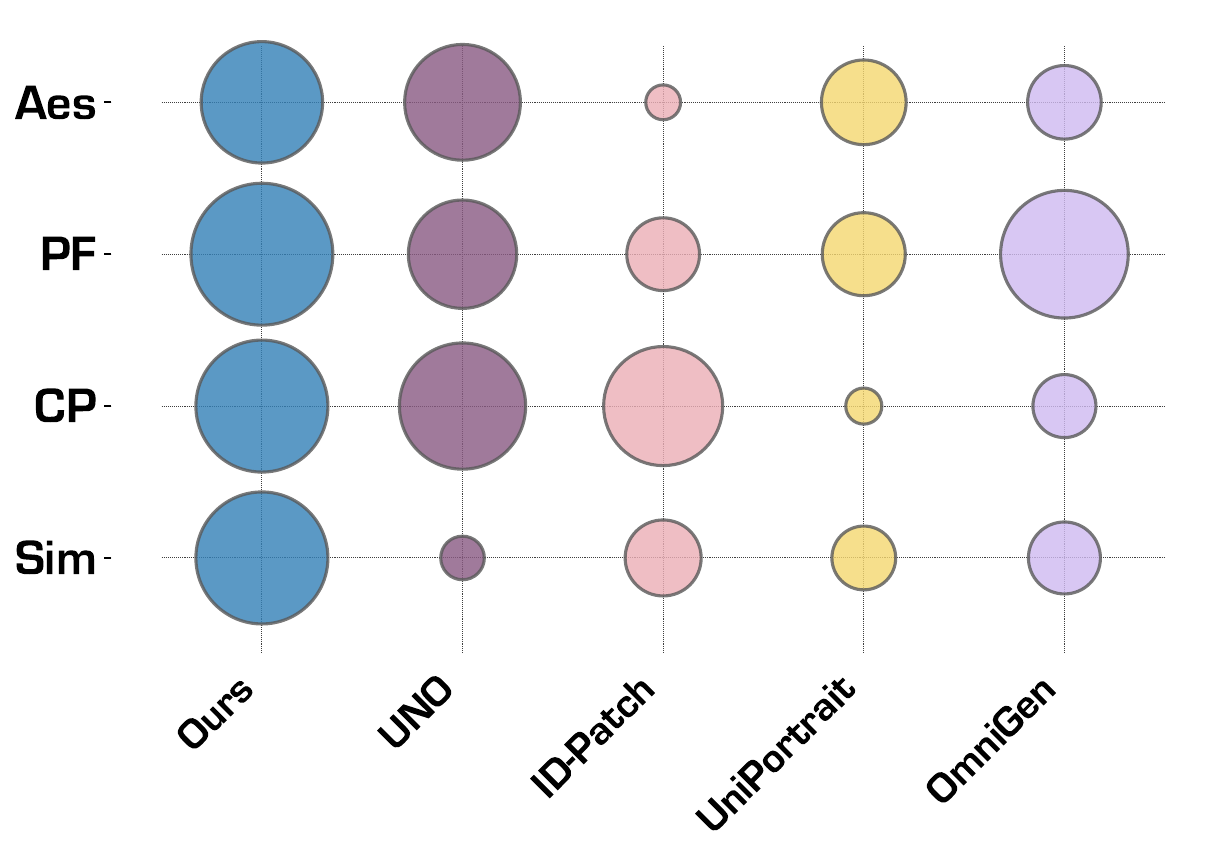}
        \captionof{figure}{\textbf{User study.} Bigger bubbles indicate higher ranking.}
        \label{fig:user_study}

\end{figure}

%% file: sections/7.conclusion.tex
\section{Conclusion} \label{sec:conclusion}

Copy-paste artifacts are a common limitation of identity customization methods, and face-similarity metrics often exacerbate the issue by implicitly rewarding direct copying. In this work, we identify and formally quantify this failure mode through \ourbenchmark, and propose targeted solutions. We curate \ourdataset~and develop training strategies and loss functions that explicitly discourage trivial replication. Empirical evaluations demonstrate that \ours~significantly reduces copy-paste artifacts while maintaining and in many cases improving identity similarity, thereby breaking the long-standing trade-off between fidelity and copying. These results highlight a practical path toward more faithful, controllable, and robust identity customization.

%% file: sections/appendix.tex

\appendix

\noindent
\textbf{\LARGE Appendix}
\vspace{5ex}

\section{Family of \ours}

FLUX.1 comprises a family of models, including FLUX.1~\cite{flux2024}, FLUX.1 Kontext~\citep{batifol2025flux} and FLUX.1 Krea~\citep{fluxkrea}. Krea is a text-to-image model with improved real-person face generation, whereas Kontext is an image-editing model that excels at making targeted adjustments while preserving the rest of the image. However, as reported in Table~\ref{tab:quantitative_comparison_single}, Kontext shows limited consistency with the reference face identity.

Our method, \ours, can be seamlessly integrated into Kontext for the face customization downstream tasks like face editing. As illustrated in Fig.~\ref{fig:kontext}, \ours~effectively injects identity information from the reference images into the target image.

The overall training pipeline follows the procedure described in Sec.~\ref{sec:method}, with a single modification: the input image provided to Kontext (whose tokens are concatenated with the noisy latent at each denoising step) is set to the target image with the face region blurred.

\input{figures/kontext}

\section{\ourdataset~Construction Details}
\label{sec:appendix_dataset}
\input{sections/a.dataset}

\section{Benchmark and Metrics Details}
\label{sec:appendix_benchmark}
\input{sections/a.benchmark.tex}

\section{Galleries of \ours}
\input{figures/gallery}
We show more results of \ours~in Fig.~\ref{fig:gallery_1}, Fig.~\ref{fig:gallery_2}, and Fig.~\ref{fig:gallery_3}.

\section{Model Framework Details}
\label{sec:model_framework}

We follow prior work \citep{ye2023ipadapter,guo2024pulid} and integrate a lightweight identity adapter into the diffusion backbone. Identity embeddings are injected by cross-attention so that the base generative prior is preserved while controllable identity signals are added.

\textbf{Face embedding.} Each reference face is first encoded by ArcFace, producing a $1\times512$ identity embedding. To match the tokenized latent space of the DiT backbone, this vector is projected with a multi-layer perceptron (MLP) into $8$ tokens of dimension $3072$ (i.e., an $8\times3072$ tensor). This tokenization provides sufficient capacity for the cross-attention layers to integrate identity cues without overwhelming the generative context.

\textbf{Controllable attribute retention.} Completely suppressing copy-like behavior is not always desirable: users sometimes expect certain mid-level appearance attributes (e.g., hairstyle, accessories) to be preserved. ArcFace focuses on high-level, identity-discriminative geometry and texture cues but omits many mid-level semantic factors. To expose controllable retention of such attributes when needed, we optionally incorporate SigLIP \citep{zhai2023siglip} as a secondary encoder. SigLIP provides more semantically entangled representations, enabling selective transfer of style-relevant traits while ArcFace anchors identity fidelity. 

\textbf{Attention mask and location control.} To further improve identity disentanglement and precise localization in the generated images, an attention mask and location control mechanism are incorporated \cite{chen2024training,xu2024permutation}. Specifically, ground-truth facial bounding boxes are extracted from the training data and used to generate binary attention masks. These masks are applied to the attention layers of the backbone model, ensuring that each reference token only attends to its corresponding face region in the image, providing location control at the same time.

\textbf{Feature injection.} After each transformer block of the DiT backbone, we inject face features through a cross-attention modulation:
\begin{equation}\label{eq:cross_attention}
    H' = H + \lambda_{\text{id}} \,\text{softmax}\!\left(\frac{(H W_Q)(E W_K)^{\top}}{\sqrt{d}} + M\right) (E W_V),
\end{equation}
where $H$ denotes the current hidden tokens, $E$ the stacked face-embedding tokens, and $W_Q, W_K, W_V$ the projection matrices; $d$ is the query/key dimension, and $\lambda_{\text{id}}=1.0$ during training. When SigLIP is enabled, its tokens are processed by a parallel cross-attention with an independent scaling coefficient.

\section{Experimental Details}
\label{sec:experiment-appendix}

\subsection{Implementation Details}
\label{sec:implementation-details}
\ours~is trained on 8 NVIDIA H100 GPUs, with a batch size of $4$ on each GPU. The learning rate is set to $1e^{-4}$, and the AdamW optimizer is employed with a weight decay of $0.01$. 
The pre-training phase runs for $60k$ steps, with a fixed prompt used during the first $20k$ steps. The subsequent paired-tuning phase lasts $30k$ steps: $50\%$ of the samples use paired (reference, ground-truth) data, while the remaining $50\%$ continue reconstruction training. Finally, a quality/style tuning stage of $10k$ steps is performed with a reduced learning rate of $1\times10^{-5}$.

For the extended ID contrastive loss, the target is used as the positve sample, while other IDs from samples in the same batch serve as negative samples. With the global batch size of $32$, this yields less than a hundred negative samples. Extended negative samples are drawn from reference bank. If this ID is identified as one of the $3$k ID in the reference bank, we simply omit its own ID and draw the from other IDs. If this ID is not identified, then it makes things easier -- all the IDs in the reference bank can be used as negative samples.  

For other baseline methods, official implementations and checkpoints (or API) are used with default settings. Methods are tested on \ourbenchmark~and real-human subset of OmniContext~\citep{wu2025omnigen2}. OmniContext uses Vision-Language Models (VLMs) to evaluate the prompt-following (PF) and subject-consistency (SC) of generated images. For reproducibility, the VLM is fixed to Qwen2.5-VL~\citep{Qwen2.5-VL}. ID-Patch~\citep{zhang2025idpatch} requires pose condition, and we use the ground-truth pose for it.

Single face embedding model may induce biased evaluation on ID similarity, thus we average three de-facto face recognition models' consine similarity to compute the overall ID similarity metric, namely ArcFace~\citep{deng2019arcface}, FaceNet~\citep{schroff2015facenet}, and AdaFace~\citep{kim2022adaface}.

\subsection{More Discussion on the Quantitative Results}
\label{sec:more_discussion_quantitative_results}

The performance of GPT on our 3‑and‑4‑people subset offers a useful validation of our copy‑paste metric, as shown in Table~\ref{tab:quantitative_comparison_multi}. This subset largely comprises group photographs from TV series that GPT may have encountered during pretraining, so GPT attains unusually high identity‑similarity scores both to the ground truth (GT) and to the reference images. Actually, in one case GPT even generates an ID from the TV series that is not present in the reference images. This behaviour approximates an idealized scenario in which a model fully understands and faithfully reproduces the target identity: similarity to GT and to references are both high, and the copy‑paste measure the difference between distances to GT and to references approaches zero. These observations are consistent with our metric design and support its ability to distinguish true identity understanding from trivial copy‑and‑paste replication.

We report the experimental limit in Table~\ref{tab:quantitative_comparison_single}. If one model completely copy the reference image, $\mathrm{Sim_{GT}} = 0.521$, $\mathrm{Sim_{Ref}} = 1.0$, and copy-paste is $0.999$, which aligns with the theoretical limit $1.0$ of copy-paste. 

The prompt-following ability is measured by CLIP-I and CLIP-T in our benchmark, and is judged by VLM in OmniContext. \ours gains state-of-the-art performance in both metrics, and is ranked the highest in our user study. However, the credibility of CLIP scores and the aesthetic scores may be debated, as they are not always consistent with human perception.

\section{Ablation Study Details}
\label{sec:appendix_ablation}
\input{figures/abl_sig_infonce}

In this section, we systematically evaluate the impact of training strategy, GT-aligned ID-Loss, InfoNCE ID Loss, and our dataset construction. User study is also conducted to validate the consistency of the proposed metrics with human perception, as well as evaluate the human preference on different methods.

\textbf{SigLIP signal.} SigLIP~\citep{zhai2023siglip} signal is introduced to retain copy-paste effect when user tend to retain the features from reference images like hairstyle, accessories, etc. As shown in Fig.~\ref{fig:ablation_siglip}, increasing the SigLIP signal weight effectively amplifies the copy-paste effect while simultaneously boosting ID similarity to the reference images exactly as expected, since stronger SigLIP guidance enforces tighter semantic alignment and transfers more fine-grained appearance cues (e.g., hairstyle, accessories, local textures).

\textbf{Training strategy.} We evaluate the effect of a paired-data fine-tuning stage. After an initial reconstruction training phase, we either continue training with paired (reference, ground-truth) data or keep training under the reconstruction objective for 10k steps. As shown in Table~\ref{tab:abl_cp_data}, continuing with paired data effectively reduces the copy-paste effect without compromising similarity to the ground truth.

\textbf{Dataset construction.} To validate the effectiveness of our dataset, we trained a model on FFHQ~\citep{karras2019style} using reconstruction training for the same number of steps. As shown in Table~\ref{tab:abl_cp_data}, the FFHQ-trained model performs poorly across all metrics. This likely stems from FFHQ's limited diversity and size, as it contains only ~70k face-only portrait images.

\textbf{GT-aligned ID-Loss.} We validate the GT-aligned ID-Loss with a simple experiment that visualizes predicted faces at different denoising time steps during training. As shown in Fig.~\ref{fig:gt_aligned_lmk}, at low noise levels the GT-aligned ID-Loss is substantially lower than the loss computed using prediction-aligned landmarks, indicating that aligning faces to ground-truth landmarks reduces denoising error and yields a more accurate identity assessment. At high noise levels the GT-aligned ID-Loss shows greater variance, producing stronger and more informative gradients that help the model learn identity features. 

\textbf{InfoNCE Loss.} The InfoNCE loss with extended negative samples is crucial for the convergence in the early training stage. We conduct a toy experiment with 1000 training samples, and record ID Loss curves with no InfoNCE loss, $0.1\times$ InfoNCE loss without extended negatives, and $0.1\times$ InfoNCE loss with extended negatives. As shown in Fig.~\ref{fig:ablation_infonce}, ID loss fits a lot faster with InfoNCE loss with extended negatives, demonstrating its effectiveness in accelerating training convergence. It also largely increases the ID similarity score, as shown in Table~\ref{tab:abl_cp_data}.

\input{tables/user_study_corr}

\section{User Study Details}
\label{sec:user_study}

\input{figures/user_study_interface}

Our user study is conducted with the same data samples and generated results in our quantitative experiments. Due to a tight financial budget, we randomly select $100$ samples from single-person subset, $100$ samples from 2-people subset, and all samples from 3-and-4 people subset. 10 participants are recruited for the study, all of whom are trained with a brief tutorial to understand the task and evaluation criteria. 

We illustrate the interface used in our user study in Fig.~\ref{fig:user_study_inferface}.


\subsection{Correlation Analysis}
\label{subsec:correlation_analysis}
We analyze the correlation between our proposed metrics and user study results. As shown in Table~\ref{tab:correlation_results}, our copy-paste metric shows a moderate positve correlation with user ratings on copy-paste effect.

\subsection{Participant Instructions}
\label{subsec:participant_instructions}

We provide the instructions for training the participants in the following table.

\begin{table*}[htpb]
\label{tab:user_study_instructions}
\centering
\begin{tcolorbox}[title=Participant Instructions and Evaluation Procedure,
enhanced,
skin first=enhanced,
skin middle=enhanced,
skin last=enhanced,
colback={rgb,255:red, 249; green, 250; blue, 255},
colframe={rgb,255:red, 109; green, 153; blue, 255},
fonttitle=\bfseries]{}

\textbf{Data source and task overview.}

Five different methods generated images under the following conditions:
\begin{itemize} 
  \item A single prompt that describes the ``ground truth image.''
  \item Between 1 and 4 people in the scene (most examples contain 1--2 people).
\end{itemize}

For each trial you will be shown the ground truth image, input images, and a generation instruction. Then you will observe five generated group-photo results (one per method) and rank them according to several evaluation dimensions. Use a 5‑star scale where 5 stars = best and 1 star = worst. Please read the input image(s) and the editing instruction carefully before inspecting the generated results.

\vspace{6pt}
\textbf{Evaluation procedure (per-image ranking).}

Rank each generated image individually on the following criteria.

\paragraph{Identity similarity}
\begin{itemize} 
  \item How well do the person(s) in the generated image resemble the person(s) in the \emph{ground truth image}?
  \item Rank images by their resemblance to the ground truth image: the more the generated person(s) look like the original reference, the higher the rating.
  \item \textbf{Important:} When judging identity similarity, ignore factors such as image quality, rendering artifacts, or general aesthetics. Focus only on how much the person(s) resemble the original reference(s). Also, try to assess resemblance to the ground truth image as a whole, rather than comparing to any single separate ``reference person n.''
\end{itemize}

\paragraph{Copy-and-paste effect (excessive mimicry of the reference)}
\begin{itemize} 
  \item Generated images should resemble the original reference but should not be direct copies of an individual reference photo.
  \item Evaluate whether the generated person appears to be directly copied from one of the reference images. Consider changes (or lack thereof) in \textbf{expression, head pose and orientation, facial expression/demeanor, and lighting/shading}.
  \item The lower the degree of direct copying (i.e., the less it looks like a pasted replica), the better. Rank according to the amount of change observed in the person(s): more natural variation (less copy-paste) should be ranked higher.
\end{itemize}

\paragraph{Prompt following}
\begin{itemize} 
  \item Does the generated image reflect the content and constraints specified by the prompt/instruction?
  \item Rank images by prompt fidelity: the more faithfully the image follows the prompt, the higher the ranking.
\end{itemize}

\paragraph{Aesthetics}
\begin{itemize} 
  \item Judge the overall visual quality and pleasantness of the generated image (e.g., smoothness of rendering, harmonious body poses and composition).
  \item Rank images by aesthetic quality: higher perceived visual quality receives higher ratings.
\end{itemize}

\end{tcolorbox}
\end{table*}


\section{Prompts for Language Models}
\label{sec:appendix_prompts}

Large language models (LLMs) and vision-language models (VLMs) are used in various stages of our work, including dataset captioning and OmniContext evaluation. 

\subsection{Dataset Captioning}
\label{subsec:dataset_captioning}

Besides the system prompt, we design 6 different prompts to generate diverse captions for each image. 1 prompt is randomly selected for each image during captioning. 

\begin{table*}[htpb]
\centering
\begin{tcolorbox}[title=Full Prompts for Dataset Captioning (6 variants),
enhanced,
skin first=enhanced,
skin middle=enhanced,
skin last=enhanced,
colback={rgb,255:red, 249; green, 250; blue, 255},
colframe={rgb,255:red, 109; green, 153; blue, 255},
fonttitle=\bfseries]{}
\textbf{System Prompt}: You are an advanced vision-language model tasked with generating accurate and comprehensive captions for images.\\

\textbf{Prompt 1:} Please provide a brief description of the image based on these guidelines:
\begin{enumerate}
  \item Describe the clothing, accessories, or jewelry worn by the people in detail.
  \item Describe the genders, actions, and posture of the individual in detail, focusing on what they are doing.
  \item The description should be concise, with a maximum of 77 words.
  \item Start with `This image shows'
\end{enumerate}

\textbf{Prompt 2:} Offer a short description of the image according to these rules:
\begin{enumerate}
  \item Focus on details about clothing, accessories, or jewelry.
  \item Focus on the gender, activity, and pose, and explain what the people is doing.
  \item Keep the description within 77 words.
  \item Begin the description with `This image shows'
\end{enumerate}

\textbf{Prompt 3:} Please describe the image briefly, following these instructions:
\begin{enumerate}
  \item Provide a detailed description of the clothing or jewelry the person may be wearing.
  \item Provide a detailed description of the two persons' gender, actions, and body position.
  \item Limit the description to no more than 77 words.
  \item Begin your description with `This image shows'
\end{enumerate}

\textbf{Prompt 4:} Describe the picture briefly according to these rules:
\begin{enumerate}
  \item Provide a detailed description of the clothing, jewelry, or accessories of the individuals.
  \item Focus on the two persons' gender, what they are doing, and their posture.
  \item Keep the description concise, within a limit of 77 words.
  \item Start your description with `This image shows'
\end{enumerate}

\textbf{Prompt 5:} Provide a short and precise description of the image based on the following guidelines:
\begin{enumerate}
  \item Describe what the person is wearing or any accessories.
  \item Focus on the gender, activities, and body posture of the person.
  \item Ensure the description is no longer than 77 words.
  \item Begin with `This image shows'
\end{enumerate}

\textbf{Prompt 6:} Briefly describe the image according to these instructions:
\begin{enumerate}
  \item Provide a precise description of the clothing, jewelry, or other adornments of the people.
  \item Focus on the person's gender, what they are doing, and their posture.
  \item The description should not exceed 77 words.
  \item Start with the phrase `This image shows'
\end{enumerate}

\end{tcolorbox}
\end{table*}

\begin{table*}[htpb]
\centering
\begin{tcolorbox}[title=Modified Prompt for OmniContext Evaluation (Face Identity Focus),
enhanced,
skin first=enhanced,
skin middle=enhanced,
skin last=enhanced,
colback={rgb,255:red, 249; green, 250; blue, 255},
colframe={rgb,255:red, 109; green, 153; blue, 255},
fonttitle=\bfseries]{}
Rate from 0 to 10: \\

\textbf{Task:} Evaluate how well the facial features in the final image match those of the individuals in the original reference images, as described in the instruction. Focus strictly on facial identity similarity; ignore hairstyle, clothing, body shape, background, and pose.

\vspace{6pt}
\textbf{Scoring Criteria}
\begin{itemize} 
  \item \textbf{0:} The facial features are \emph{completely different} from those in the reference images.
  \item \textbf{1--3:} The facial features have \emph{minimal similarity} with only one or two matching elements.
  \item \textbf{4--6:} The facial features have \emph{moderate similarity} but several important differences remain.
  \item \textbf{7--9:} The facial features are \emph{highly similar} with only minor discrepancies.
  \item \textbf{10:} The facial features are \emph{perfectly matched} to those in the reference images.
\end{itemize}

\vspace{6pt}
\textbf{Pay detailed attention to these facial elements:}
\begin{itemize} 
  \item \textbf{Eyes:} Shape, size, spacing, color, and distinctive characteristics of the eyes and eyebrows.
  \item \textbf{Nose:} Shape, size, width, bridge height, and nostril appearance.
  \item \textbf{Mouth:} Lip shape, fullness, width, and distinctive smile characteristics.
  \item \textbf{Facial structure:} Cheekbone prominence, jawline definition, chin shape, and forehead structure.
  \item \textbf{Skin features:} Distinctive marks like moles, freckles, wrinkles, and overall facial texture.
  \item \textbf{Proportions:} Overall facial symmetry and proportional relationships between features.
\end{itemize}

\vspace{6pt}
\textbf{Example:} If the instruction requests combining the face from one image onto another pose, the final image should clearly show the \emph{same} facial features from the source image.

\vspace{6pt}
\textbf{Important:}
\begin{itemize} 
  \item For each significant facial feature difference, deduct at least one point.
  \item \emph{Ignore} hairstyle, body shape, clothing, background, pose, or other non-facial elements.
  \item Focus \emph{only} on facial similarity, not whether the overall instruction was followed.
  \item \textbf{Scoring should be strict} high scores should only be given for very close facial matches.
  \item Consider the level of detail visible in the images when making your assessment.
\end{itemize}

\vspace{6pt}
\textbf{Editing instruction:} \texttt{\textless instruction\textgreater}
\end{tcolorbox}
\end{table*}

%% file: figures/kontext.tex
\begin{figure*}[b]
    \centering
    \includegraphics[width=\linewidth]{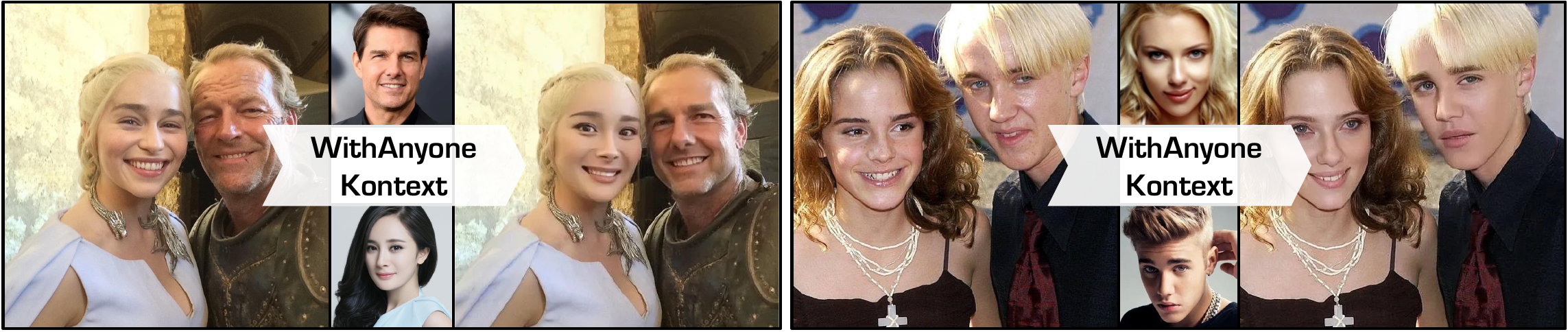}
    \caption{\small \textbf{Application of \ours-Kontext.} Marrying editing models, \ours~is capable of face editing given customization references.}
    \label{fig:kontext}
\end{figure*}

%% file: sections/a.dataset.tex
To fill in the void left by the lack of publicly available multi-ID datasets, a data constraction pipeline is proposed to create a large-scale dataset of multi-person images with paired identity references for identities on the data record. Based on this pipeline, $500$k group photo images are collected, featuring $3$k identities, each with hundreds of single-ID reference images. Another $1$M images that cannot be identified are also included in the dataset for image reconstruction training purpose for image reconstruction training purpose. 

\subsection{Dataset Construction Pipeline}\label{subsec:dataset_construction_pipeline}

The pipeline contains four steps, as shown in Fig.~\ref{fig:pipeline}. The detailed pipeline are as follows.

\input{figures/data_stats}

\textbf{Single-ID images.} To construct a ID reference set, single-ID images were collected from the web using celebrity names as search queries on Google Images. For each image, facial features were extracted with ArcFace~\citep{ren2023pbidr}, ensuring that only images containing exactly one face were retained. To remove outliers, DBSCAN~\citep{schubert2017dbscan} clustering was applied to the embeddings for each celebrity, resulting in a set of cluster centers and hundreds of reference images per identity. This process established a reliable reference set for each unique identity. Human review confirms the accuracy of the ID bank built in this step.

\textbf{Multi-ID images}. To achieve best searching efficiency, group photos were obtained using more complex queries that combined multiple celebrity names, keywords indicating the number of people (e.g., ``two celebrities''), scene descriptors (e.g., ``award ceremony''), and negative keywords to filter out irrelevant results. ArcFace embeddings were extracted for these images, yielding a large pool of candidate multi-ID images. At this stage, the dataset comprised more than 20 million images.

\textbf{Retrieval.} To provide ID reference for the multi-ID images, it is necessary to retrieve the IDs on it. All single-ID cluster centers were aggregated into an embedding matrix. For each detected face in every multi-ID image, its ArcFace embedding was compared to all single-ID cluster centers to determine identity. The similarity between two embeddings was calculated as:
\begin{equation}\label{eq:arcface_sim}
    \mathrm{sim}(id_1, id_2) = \cos(f(id_{1}), f(id_{2}))
\end{equation}
where $id_{1}$ and $id_{2}$ denote two faces, and $f$ is the ArcFace embedding network.

Each face in a multi-ID image was assigned the identity of the single-ID cluster center with the highest similarity, provided the similarity exceeded a predefined threshold (0.5). This approach enabled accurate and automated identity assignment in group images and facilitated retrieval of corresponding reference images.

\textbf{Filtering and labelling.} To further improve dataset quality, a series of annotation and filtering steps were applied. The Recognize Anything model~\citep{zhang2023recognize}, an aesthetic score predictor~\citep{aesthetic-predictor-v2-5}, and other auxiliary tools were used for annotation. Images with low aesthetic scores or those identified as collages rather than genuine group photos were excluded. Optical Character Recognition (OCR) tools detected watermarks and logos, which were cropped out when possible; otherwise, the images were discarded. Finally, descriptive captions were generated for the images using a large language model, enriching the dataset with textual information.


So far, a dataset with three parts is obtained: (1) $1$M single-ID images as reference bank, or single-ID cross-paired training; (2) $500$k paired multi-ID images with identified persons; (3) $1$M unpaired multi-ID images, which can be used for training scenario without the need of references, such as reconstruction.

\input{figures/dataset_action_clothes}

\subsection{Dataset Statistics}\label{subsec:dataset_statistics}

Following prior arts~\cite{liu2015faceattributes,cheng2022generalizable,cheng2023dna,pan2023renderme}, comprehensive statistics of the dataset are provided in Fig.~\ref{fig:data_stats}, including the distribution of nationalities, the count of appearances per identity, and a word cloud illustrating the most frequent terms in the generated image captions, offering insights into the diversity and richness of the dataset. 
A long-tail distribution is observed in the count of appearances per identity in Fig.~\ref{fig:ID}, with a few identities appearing frequently while many others are less common. This provide a diverse set of identities, as well as a perfect test dataset without identity interaction with the training set. Fig.~\ref{fig:nation} and Fig.~\ref{fig:word_cloud} illustrate \ourdataset's nationality distribution and action diversity respectively.
The comparison between the proposed dataset and existing multi-ID datasets are listed in Table~\ref{tab:dataset_comparison}, highlighting \ourdataset's outstanding volume and paired references.

\input{tables/dataset_review.tex}

%% file: figures/data_stats.tex
\captionsetup[sub]{labelformat=simple}
\begin{figure*}[t]
    \centering
    \begin{subfigure}[b]{0.302\linewidth}
        \includegraphics[width=\linewidth]{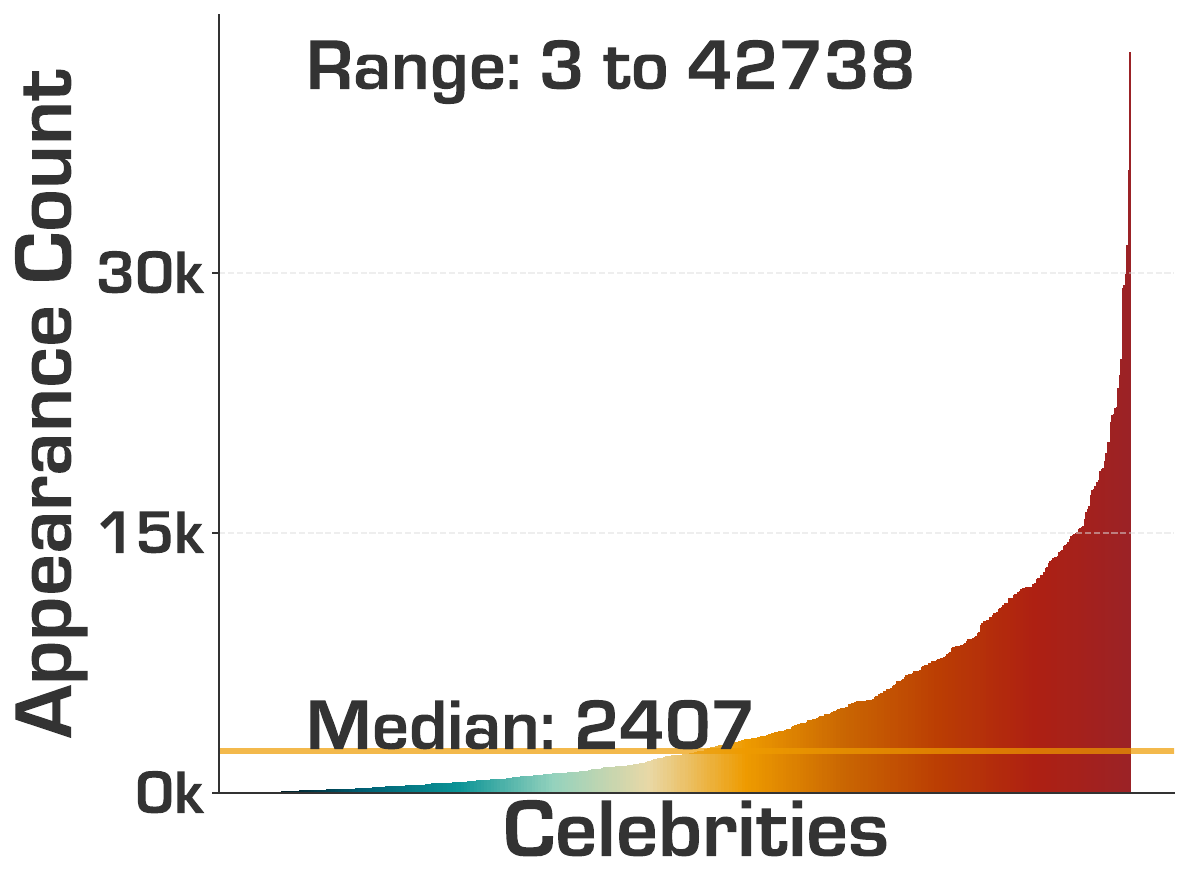}
        \caption{\small \textbf{ID Appearance.}}
        \label{fig:ID}
    \end{subfigure}
    \hspace{1ex}
    \begin{subfigure}[b]{0.334\linewidth}
        \includegraphics[width=\linewidth]{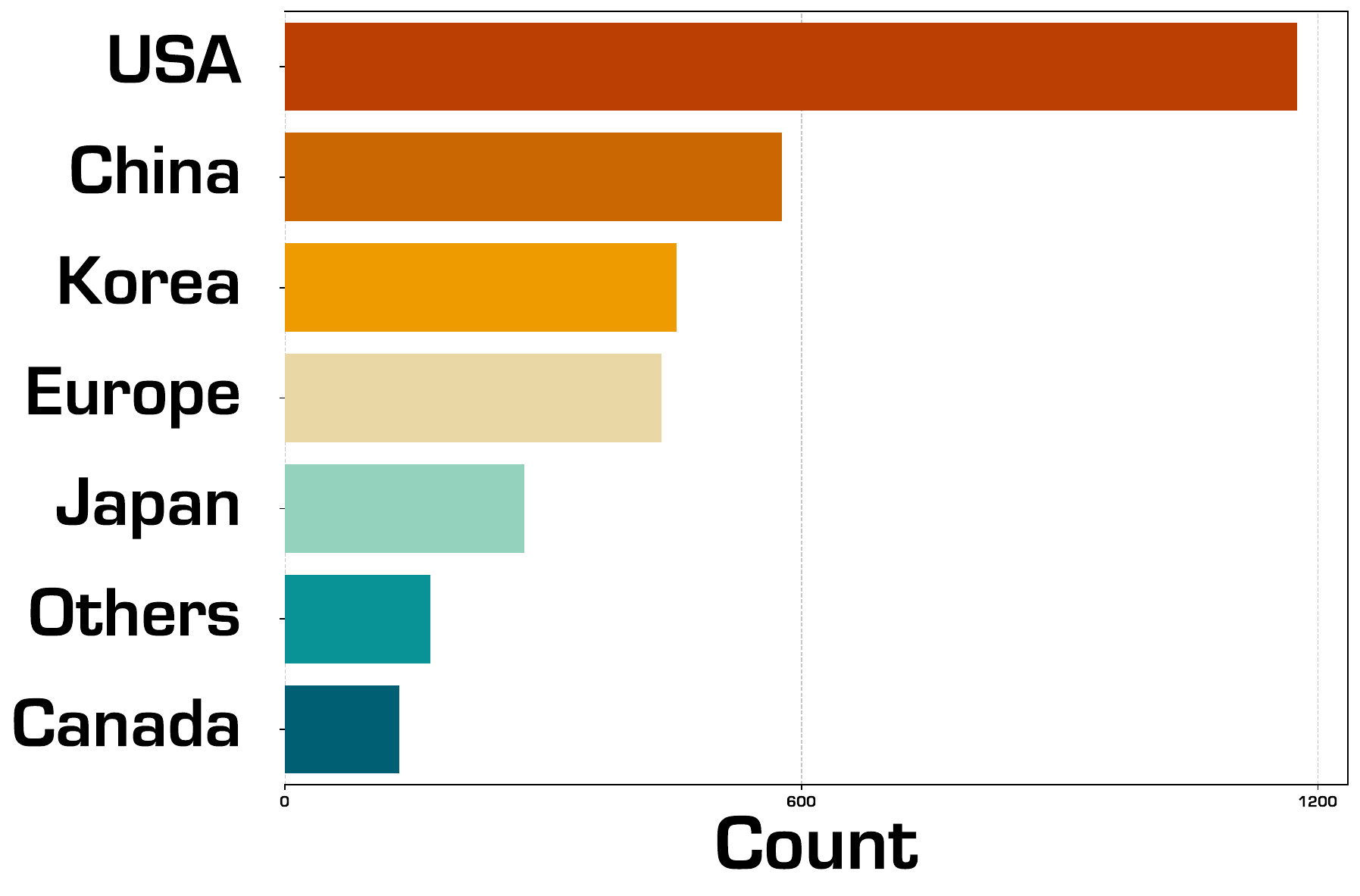}
        \caption{\small \textbf{Nationality Distribution.}}
        \label{fig:nation}
    \end{subfigure}
    \hspace{2ex}
    \begin{subfigure}[b]{0.275\linewidth}
        \includegraphics[width=\linewidth]{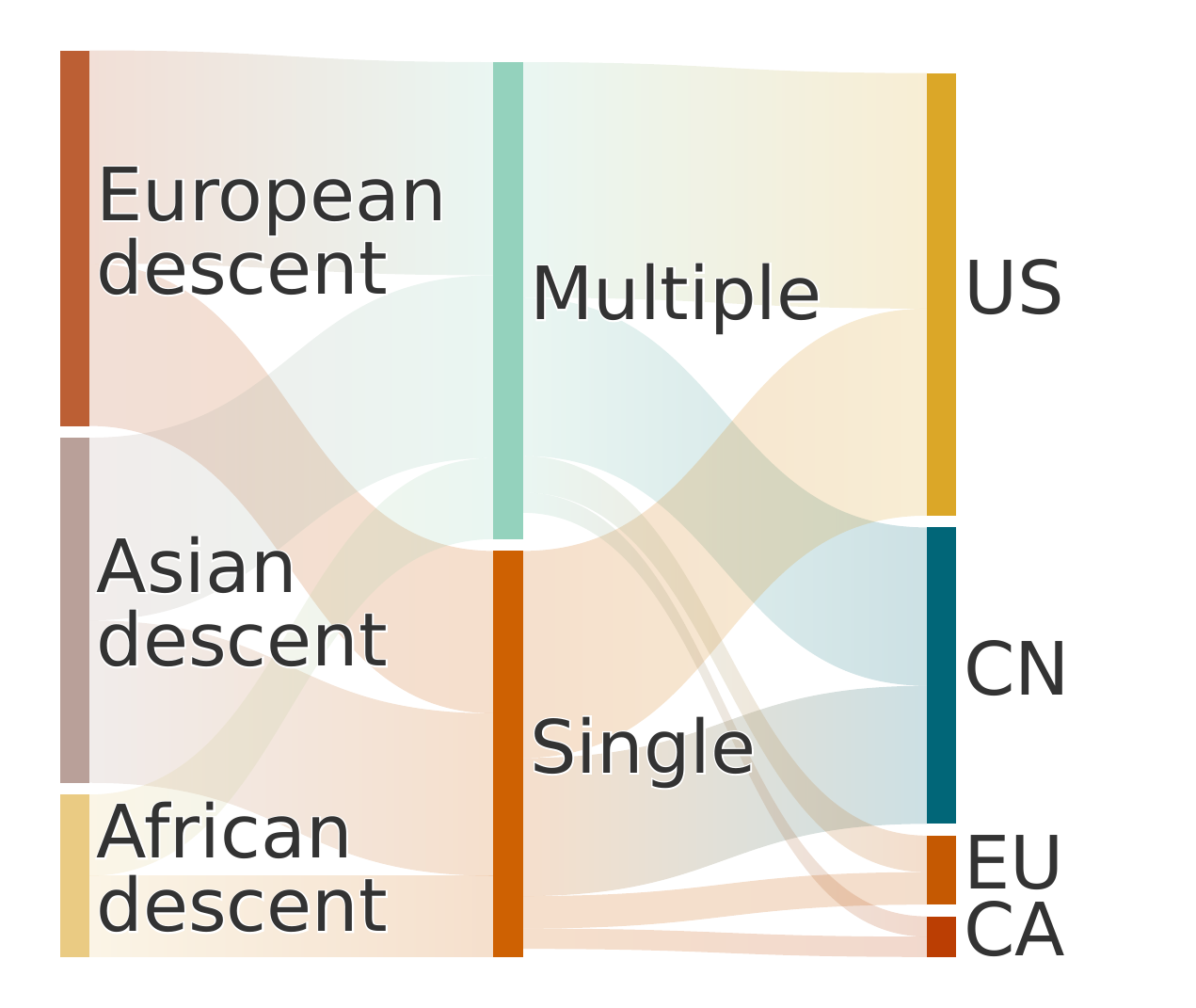}
        \caption{\small \textbf{Benchmark Distribution.}}
        \label{fig:word_cloud}
    \end{subfigure}

    \setlength{\abovecaptionskip}{2pt}
    \caption{\small \textbf{Overview of Dataset Distributions.} (a) ID appearance distribution for the subset of one nation: the x-axis represents celebrities, sorted by the number of images in which they appear.   (b) Nationality distribution: celebrities in our dataset come from over 10 countries, with most data sourced from China and the USA. (c) Word cloud of the most frequent words in the captions.}
    \label{fig:data_stats}
\end{figure*}

%% file: figures/dataset_action_clothes.tex
\captionsetup[sub]{labelformat=simple}
\begin{figure*}[h]
    \centering
    \begin{subfigure}[b]{0.48\linewidth}
        \includegraphics[width=\linewidth]{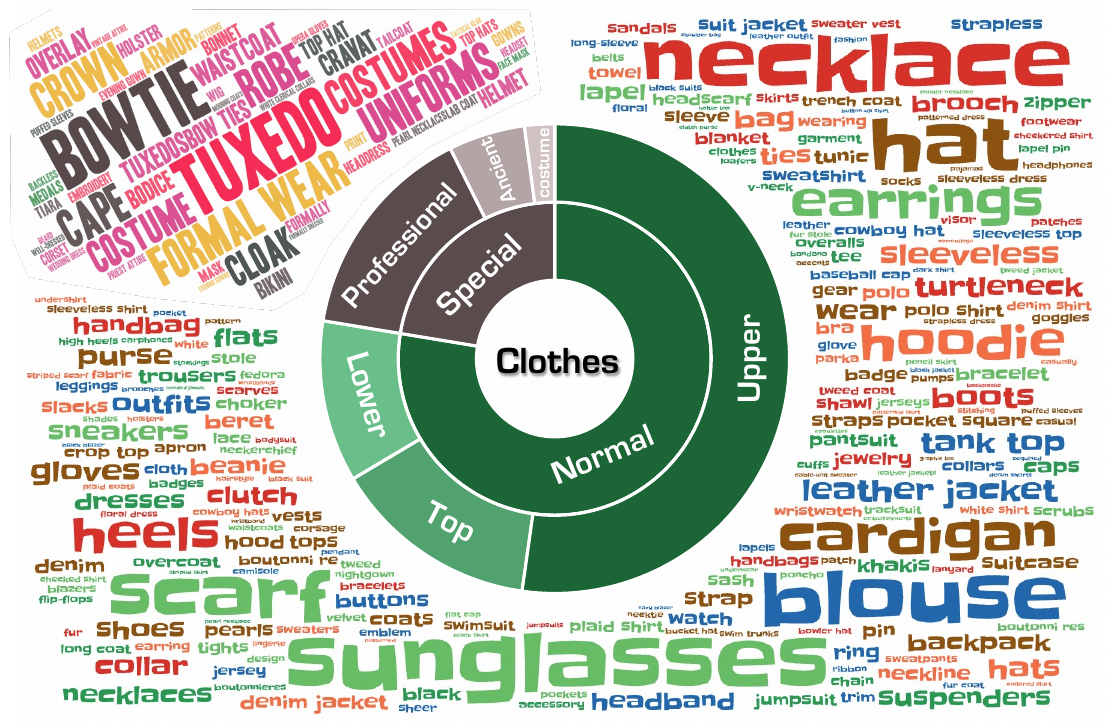}
        \caption{\small \textbf{Clothes \& Accessories Distribution.}}
        \label{fig:ID}
    \end{subfigure}
    \hspace{2ex}
    \begin{subfigure}[b]{0.48\linewidth}
        \includegraphics[width=\linewidth]{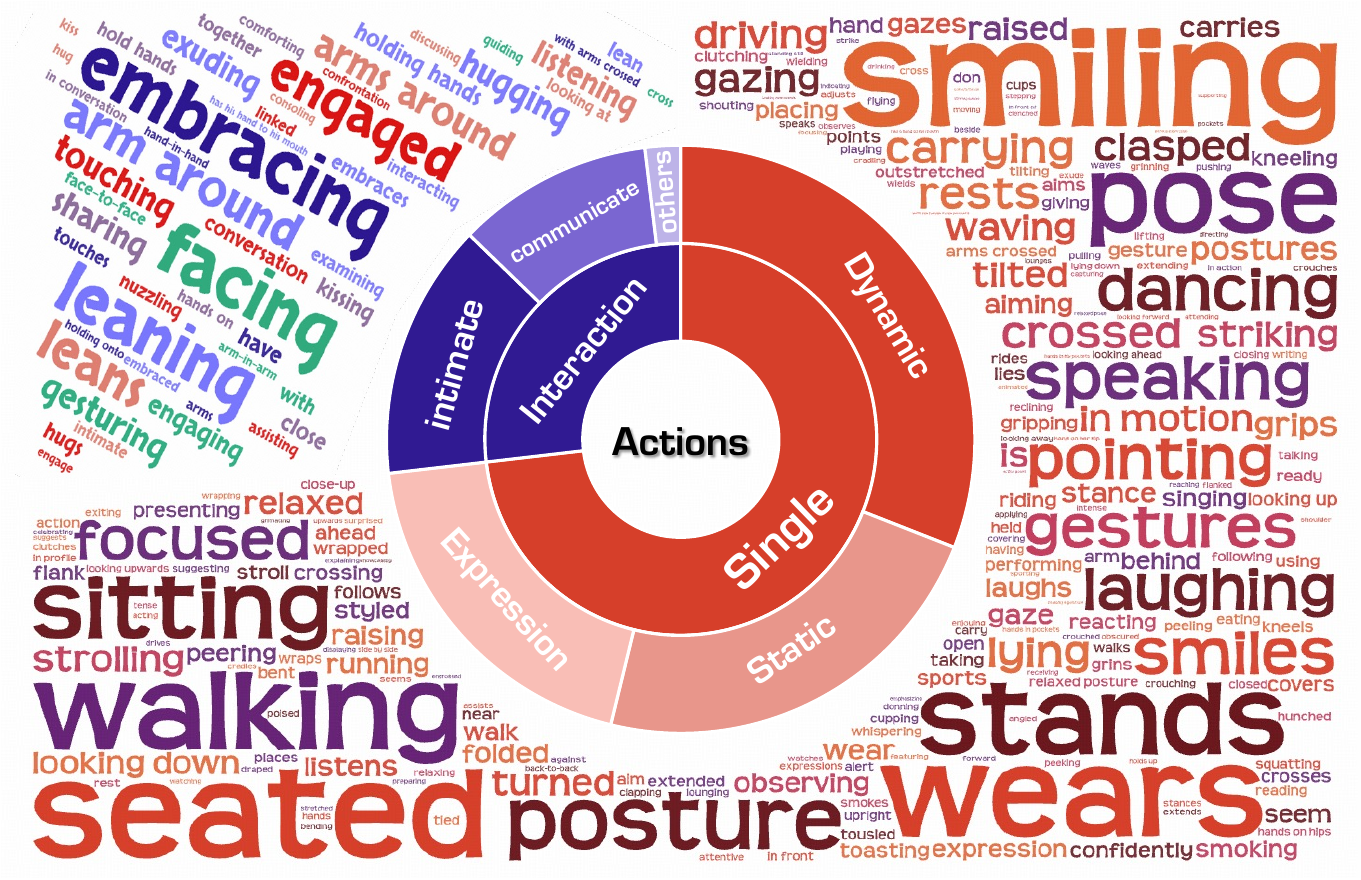}
        \caption{\small\textbf{Action Distribution.}}
        \label{fig:nation}
    \end{subfigure}

    \setlength{\abovecaptionskip}{2pt}
    \caption{\textbf{Distribution of Clothes and Action Labels of Proposed Dataset.}}
    \label{fig:data_stats}
\end{figure*}

%% file: tables/dataset_review.tex
\begin{table}[htbp]
  \centering
  \scriptsize 

  \caption{\textbf{Statistic comparison for multi-identity group photo datasets}. \textbf{\#Img} refers to total scale of the dataset; \textbf{\#Paired} refers to paired group image number; \textbf{\#Img / ID} indicates number of reference image for each single ID; \textbf{\#ID / Img} means number of IDs appears on group photos.}
  \label{tab:dataset_comparison}

  \begin{tabularx}{\linewidth}{c|c|c|c|c}
    \toprule
    \textbf{Dataset} & \textbf{\#Img} & \textbf{\#Paired}  & \textbf{\#Img / ID} & \textbf{\#ID / Img}\\
    \midrule
    IMAGO~\citep{stacchio2020imago} & $80$k & $0$  & $0$ & - \\
    MHP~\citep{chu2024uniparser} & $5$k & $0$ & $0$ & $2-10$ \\
    PIPA~\citep{zhang2015beyond} & $40$k & $40$k  & cross & $1-10$ \\
    HumanRef~\citep{jiang2025referringperson} & $36$k & $36$ & $1+$ & $1-14+$\\
    Celebrity Together~\citep{zhong2018compact} & $194$k & $0$  & $0$ & $1-5$ \\
    \midrule
    \textbf{\ourdataset} & \textbf{$1.5$M} & \textbf{$500$k}  & \textbf{$100+$} & $1-5$ \\
    \bottomrule
  \end{tabularx}

\end{table}

%% file: sections/a.benchmark.tex
Most existing methods are evaluated on privately curated test sets that are seldom released, and even when datasets are shared, the accompanying evaluation protocols vary widely. For example, ID-Patch \citep{zhang2025idpatch} and UniPortrait \citep{he2024uniportrait} measure identity similarity using ArcFace embeddings, whereas UNO \citep{wu2025uno} relies on DINO \citep{oquab2023dinov2} and CLIP similarity scores. This heterogeneity together with the common practice of reporting only the cosine similarity between matched ArcFace embeddings fails to capture more nuanced insights and can even encourage degenerate behavior in which models produce images that are effectively “copy‑pastes” of the reference photos.

In this work, \ourbenchmark~ is introduced as a unified and extensible evaluation framework for group photo (multi-ID) generation. It standardizes assessment along two complementary axes: (i) identity fidelity (preserving each target identity without unintended copying and blending), and (ii) generation quality (semantic faithfulness to the prompt/ground truth and overall aesthetic quality).

The data used in \ourbenchmark~ are drawn from the long-tail portion of \ourdataset. We first select the least frequent identities and gather all images containing them. To prevent information leakage, the training split is filtered to ensure zero identity overlap with the benchmark set. The final benchmark contains 435 samples; each sample provides 1--4 reference identities (with their images), a corresponding ground-truth image, and a text prompt describing that ground-truth scene.

\textbf{Identity Blending.} In the similarity matrix, the off-diagonal elements correspond to the similarity between different identities. The average of the diagonal elements is used as the metric for identity fidelity, and the average of the off-diagonal elements serves as the metric for identity blending, as in Eq.~\ref{eq:identity_blending}. 

\begin{equation}\label{eq:identity_blending}
    \mathrm{M_{Bld}}(x^{g}, x^{t}) = \frac{1}{N^2 - N} \sum_{i=1}^{N} \sum_{j=1, j \neq i}^{N} \cos(g_{i}, t_{j})
\end{equation}
where $g_i$ is the embedding of the $i$-th face in the generated image $x^g$, and $t_j$ is the embedding of the $j$-th face in the ground-truth image $x^t$. A lower value indicates less unintended blending between different identities, which is desirable.

\textbf{Generation quality.} The overall generation quality is evaluated based on CLIP-I and CLIP-T, which are the de facto standards for evaluating the prompt-following capability \citep{radford2021clip}, are employed to measure the cosine similarity in the CLIP embedding space between the generated image and the ground truth image or caption. Additionally, an aesthetic score model \citep{aesthetic-predictor-v2-5} is used to assess the aesthetic quality of the generated images.

%% file: figures/gallery.tex
\begin{figure*}[h]
    \centering
    \includegraphics[width=0.9\textwidth]{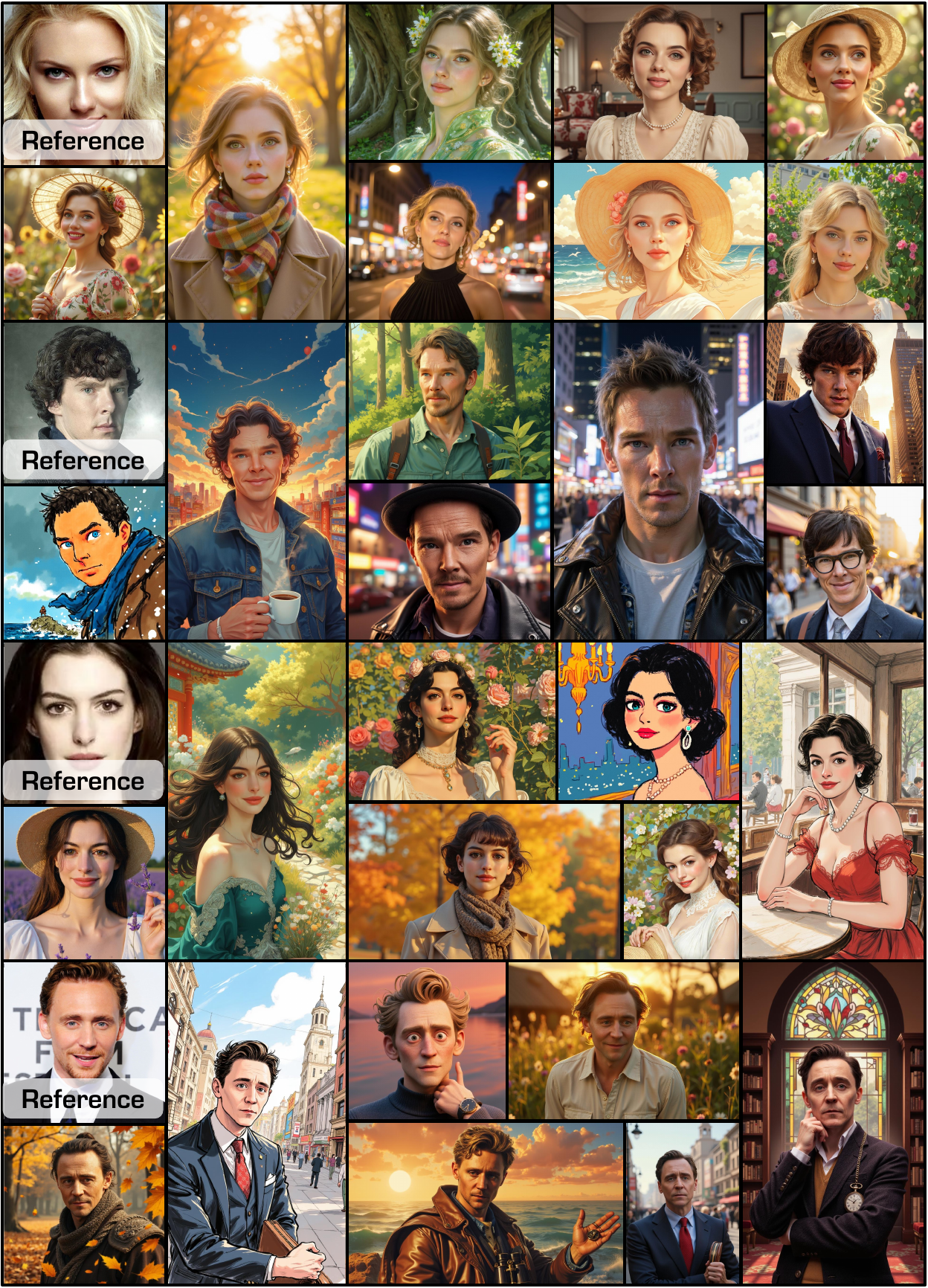}
    \caption{\small \textbf{Galleries of Single-ID Generation}.}
    \label{fig:gallery_1}
\end{figure*}

\begin{figure*}[h]
    \centering
    \includegraphics[width=0.75\textwidth]{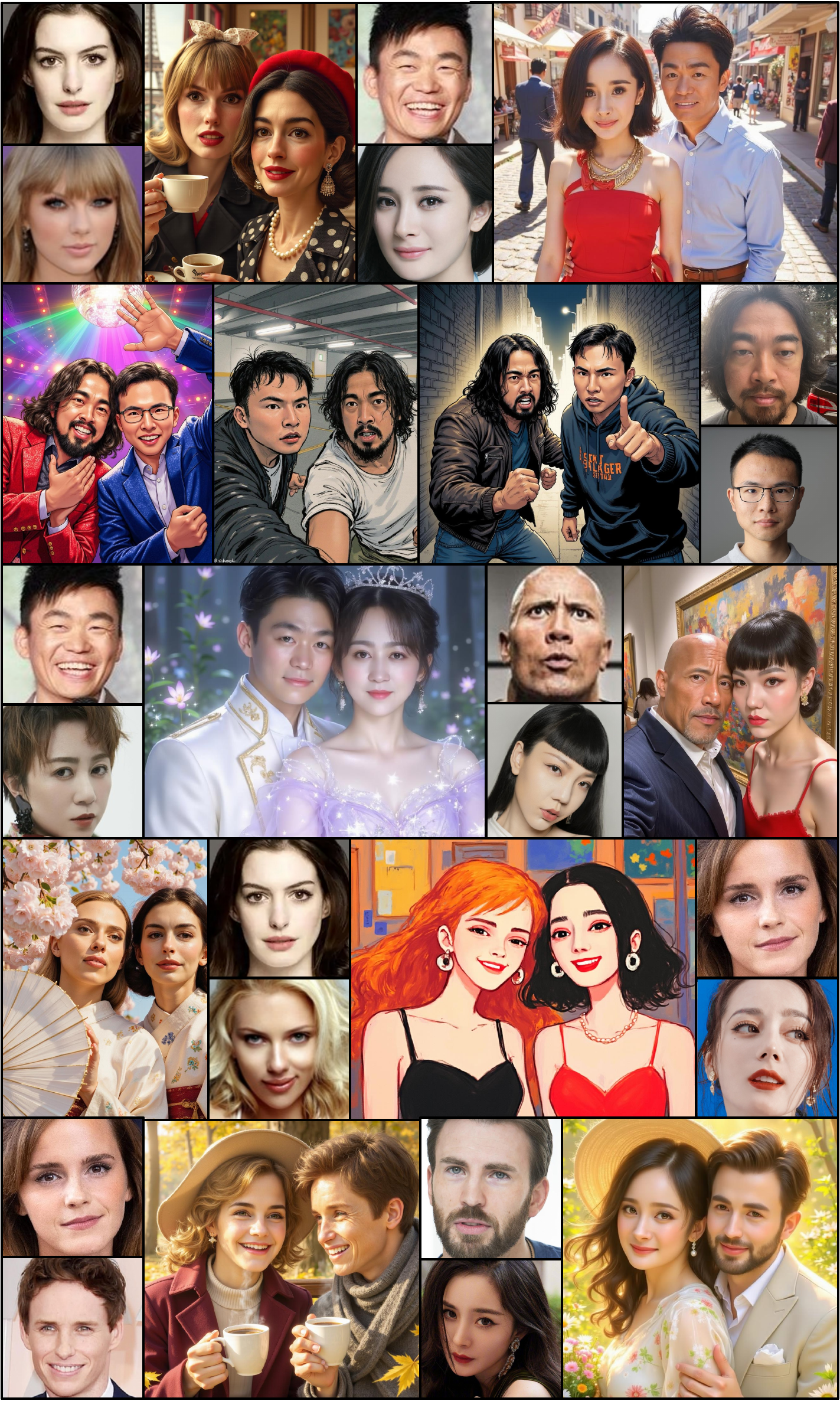}
    \caption{\small \textbf{Galleries of 2-person Generation}.}
    \label{fig:gallery_2}
\end{figure*}

\begin{figure*}[h]
    \centering
    \includegraphics[width=0.75\textwidth]{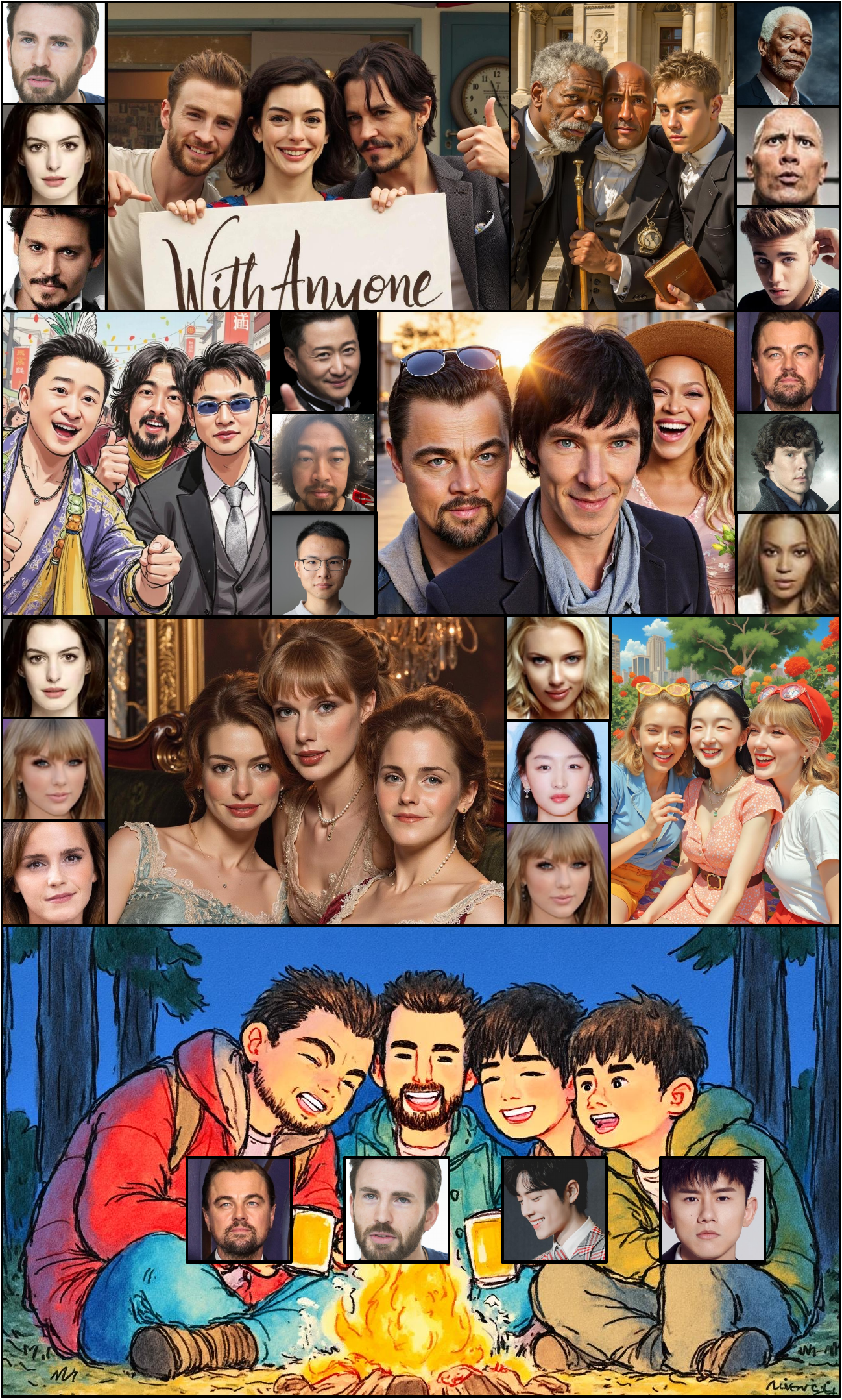}
    \caption{\small \textbf{Galleries of 3-to-4-person Generation}.}
    \label{fig:gallery_3}
\end{figure*}

%% file: figures/abl_sig_infonce.tex
\begin{figure*}[t]
    \centering
    \begin{minipage}{0.48\linewidth}
        \centering
        \includegraphics[width=\linewidth]{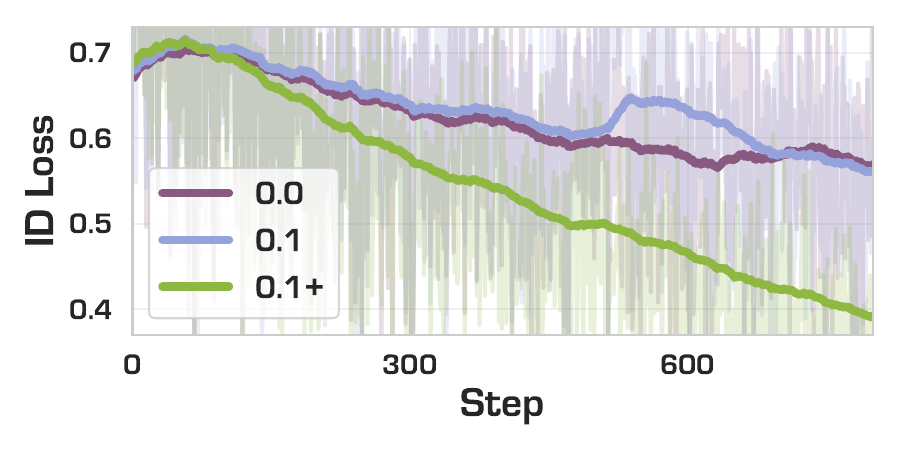}
        \caption{\small \textbf{ID Loss Curves with $\lambda \times$ InfoNCE Loss.} $0.1$ is $0.1 \times$ InfoNCE Loss without extended negative samples, and $0.1+$ is $0.1 \times$ InfoNCE Loss with extended negative samples.}
        \label{fig:ablation_infonce}
    \end{minipage}\hfill
    \begin{minipage}{0.48\linewidth}
        \centering
        \includegraphics[width=\linewidth]{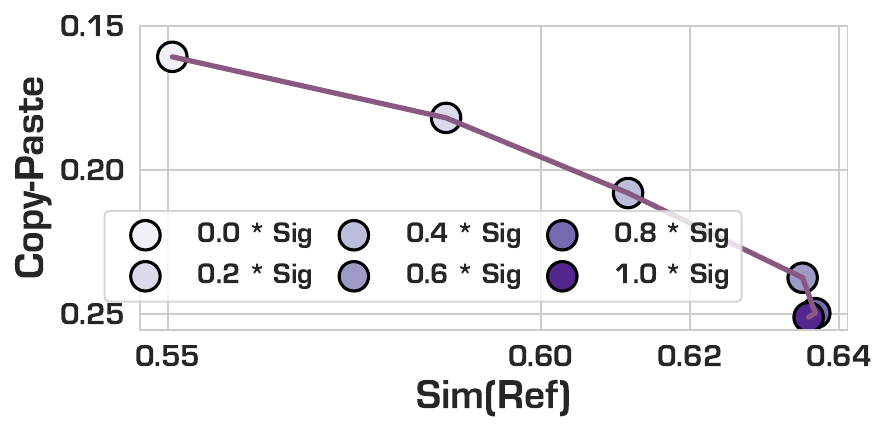}
        \caption{\small \textbf{Trade-off Curves} with $\lambda \times$ Siglip and $(1-\lambda)\times$ ArcFace signal.}
        \label{fig:ablation_siglip}
    \end{minipage}
\end{figure*}

%% file: tables/user_study_corr.tex
\begin{table*}[b]
  \centering
  \small
  \renewcommand{\arraystretch}{1}
  \caption{\small \textbf{Correlation Statistics Between Machine Ranking and Human Ranking.} Reported values include Pearson’s $r$, Spearman’s $\rho$, and Kendall’s $\tau$ with corresponding $p$-values.}
  \label{tab:correlation_results}
  \begin{tabular}{l|c|c|c}
    \toprule
    \textbf{Dimension (N)} & \textbf{Pearson $r$ (p)} & \textbf{Spearman $\rho$ (p)} & \textbf{Kendall $\tau$ (p)} \\
    \midrule
    Copy-Paste & $0.4417$ ($7.98e{-48}$) & $0.4535$ ($1.26e{-50}$) & $0.3405$ ($1.10e{-46}$) \\
    ID Sim & $0.3254$ ($1.54e{-26}$) & $0.3237$ ($2.91e{-26}$) & $0.2423$ ($1.11e{-25}$) \\
    \bottomrule
  \end{tabular}
\end{table*}

%% file: figures/user_study_interface.tex
\begin{figure}
    \centering
    \includegraphics[width=1\linewidth]{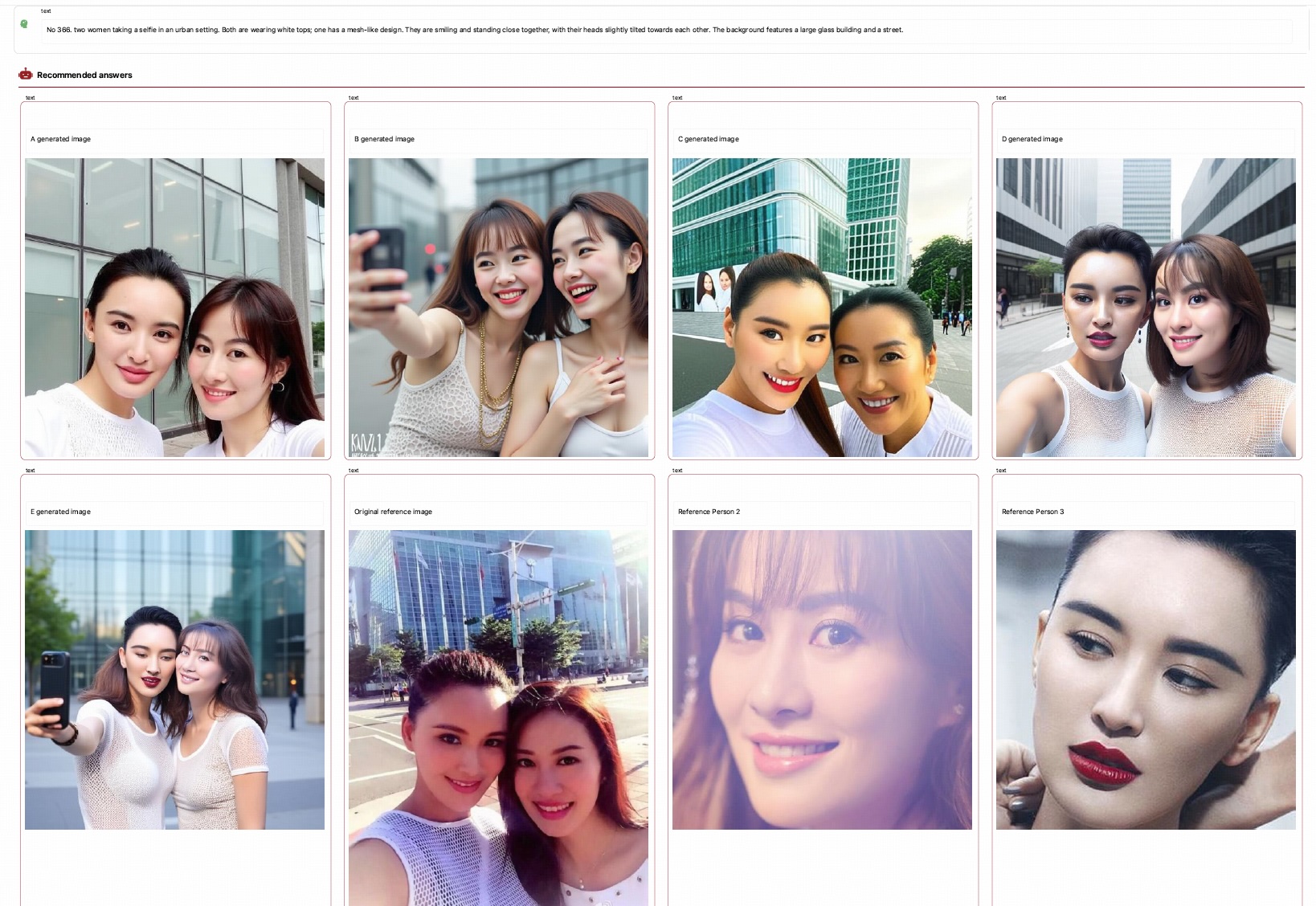}
    \caption{\small \textbf{User Study Interface}. }
    \label{fig:user_study_inferface}
\end{figure}

%% file: main.bbl
\begin{thebibliography}{10}

\bibitem{Qwen2.5-VL}
Shuai Bai, Keqin Chen, Xuejing Liu, Jialin Wang, Wenbin Ge, Sibo Song, Kai Dang, Peng Wang, Shijie Wang, Jun Tang, Humen Zhong, Yuanzhi Zhu, Mingkun Yang, Zhaohai Li, Jianqiang Wan, Pengfei Wang, Wei Ding, Zheren Fu, Yiheng Xu, Jiabo Ye, Xi~Zhang, Tianbao Xie, Zesen Cheng, Hang Zhang, Zhibo Yang, Haiyang Xu, and Junyang Lin.
\newblock Qwen2.5-vl technical report.
\newblock {\em arXiv preprint arXiv:2502.13923}, 2025.

\bibitem{batifol2025flux}
Stephen Batifol, Andreas Blattmann, Frederic Boesel, Saksham Consul, Cyril Diagne, Tim Dockhorn, Jack English, Zion English, Patrick Esser, Sumith Kulal, et~al.
\newblock Flux. 1 kontext: Flow matching for in-context image generation and editing in latent space.
\newblock {\em arXiv e-prints}, 2025.

\bibitem{chen2024training}
Anthony Chen, Jianjin Xu, Wenzhao Zheng, Gaole Dai, Yida Wang, Renrui Zhang, Haofan Wang, and Shanghang Zhang.
\newblock Training-free regional prompting for diffusion transformers.
\newblock {\em arXiv preprint arXiv:2411.02395}, 2024.

\bibitem{chen2025xverse}
Bowen Chen, Mengyi Zhao, Haomiao Sun, Li~Chen, Xu~Wang, Kang Du, and Xinglong Wu.
\newblock Xverse: Consistent multi-subject control of identity and semantic attributes via dit modulation.
\newblock {\em arXiv preprint arXiv:2506.21416}, 2025.

\bibitem{chen2024id}
Weifeng Chen, Jiacheng Zhang, Jie Wu, Hefeng Wu, Xuefeng Xiao, and Liang Lin.
\newblock Id-aligner: Enhancing identity-preserving text-to-image generation with reward feedback learning.
\newblock {\em arXiv preprint arXiv:2404.15449}, 2024.

\bibitem{cheng2023dna}
Wei Cheng, Ruixiang Chen, Siming Fan, Wanqi Yin, Keyu Chen, Zhongang Cai, Jingbo Wang, Yang Gao, Zhengming Yu, Zhengyu Lin, et~al.
\newblock Dna-rendering: A diverse neural actor repository for high-fidelity human-centric rendering.
\newblock In {\em ICCV}, 2023.

\bibitem{cheng2022generalizable}
Wei Cheng, Su~Xu, Jingtan Piao, Chen Qian, Wayne Wu, Kwan-Yee Lin, and Hongsheng Li.
\newblock Generalizable neural performer: Learning robust radiance fields for human novel view synthesis.
\newblock {\em arXiv preprint arXiv:2204.11798}, 2022.

\bibitem{cheng2025umo}
Yufeng Cheng, Wenxu Wu, Shaojin Wu, Mengqi Huang, Fei Ding, and Qian He.
\newblock Umo: Scaling multi-identity consistency for image customization via matching reward.
\newblock {\em arXiv preprint arXiv:2509.06818}, 2025.

\bibitem{chu2024uniparser}
Jiaming Chu, Lei Jin, Yinglei Teng, Jianshu Li, Yunchao Wei, Zheng Wang, Junliang Xing, Shuicheng Yan, and Jian Zhao.
\newblock Uniparser: Multi-human parsing with unified correlation representation learning.
\newblock {\em TIP}, 2024.

\bibitem{Deng2020retinaface}
Jiankang Deng, Jia Guo, Evangelos Ververas, Irene Kotsia, and Stefanos Zafeiriou.
\newblock Retinaface: Single-shot multi-level face localisation in the wild.
\newblock In {\em CVPR}, 2020.

\bibitem{deng2019arcface}
Jiankang Deng, Jia Guo, Niannan Xue, and Stefanos Zafeiriou.
\newblock Arcface: Additive angular margin loss for deep face recognition.
\newblock In {\em CVPR}, 2019.

\bibitem{aesthetic-predictor-v2-5}
discus0434.
\newblock aesthetic-predictor-v2-5.
\newblock \url{https://github.com/discus0434/aesthetic-predictor-v2-5}, 2023.
\newblock Accessed: 2025-05-12.

\bibitem{esser2024scaling}
Patrick Esser, Sumith Kulal, Andreas Blattmann, Rahim Entezari, Jonas M{\"u}ller, Harry Saini, Yam Levi, Dominik Lorenz, Axel Sauer, Frederic Boesel, et~al.
\newblock Scaling rectified flow transformers for high-resolution image synthesis.
\newblock In {\em ICML}, 2024.

\bibitem{guo2024pulid}
Zinan Guo, Yanze Wu, Zhuowei Chen, Lang Chen, Peng Zhang, and Qian He.
\newblock Pulid: Pure and lightning id customization via contrastive alignment.
\newblock In {\em NeurIPS}, 2024.

\bibitem{he2024uniportrait}
Junjie He, Yifeng Geng, and Liefeng Bo.
\newblock Uniportrait: A unified framework for identity-preserving single-and multi-human image personalization.
\newblock {\em ICCV}, 2025.

\bibitem{hertz2022prompt}
Amir Hertz, Ron Mokady, Jay Tenenbaum, Kfir Aberman, Yael Pritch, and Daniel Cohen-Or.
\newblock Prompt-to-prompt image editing with cross attention control.
\newblock {\em ICLR}, 2023.

\bibitem{ho2020ddpm}
Jonathan Ho, Ajay Jain, and Pieter Abbeel.
\newblock Denoising diffusion probabilistic models.
\newblock {\em NeurIPS}, 2020.

\bibitem{hu2025dynamicid}
Xirui Hu, Jiahao Wang, Hao Chen, Weizhan Zhang, Benqi Wang, Yikun Li, and Haishun Nan.
\newblock Dynamicid: Zero-shot multi-id image personalization with flexible facial editability.
\newblock {\em ICCV}, 2025.

\bibitem{hu2025simulating}
Yuqi Hu, Longguang Wang, Xian Liu, Ling-Hao Chen, Yuwei Guo, Yukai Shi, Ce~Liu, Anyi Rao, Zeyu Wang, and Hui Xiong.
\newblock Simulating the real world: A unified survey of multimodal generative models.
\newblock {\em arXiv preprint arXiv:2503.04641}, 2025.

\bibitem{hyung2024magicapture}
Junha Hyung, Jaeyo Shin, and Jaegul Choo.
\newblock Magicapture: High-resolution multi-concept portrait customization.
\newblock In {\em AAAI}, 2024.

\bibitem{jiang2025infiniteyou}
Liming Jiang, Qing Yan, Yumin Jia, Zichuan Liu, Hao Kang, and Xin Lu.
\newblock Infiniteyou: Flexible photo recrafting while preserving your identity.
\newblock {\em ICCV}, 2025.

\bibitem{jiang2025referringperson}
Qing Jiang, Lin Wu, Zhaoyang Zeng, Tianhe Ren, Yuda Xiong, Yihao Chen, Qin Liu, and Lei Zhang.
\newblock Referring to any person.
\newblock 2025.

\bibitem{karras2017progressive}
Tero Karras, Timo Aila, Samuli Laine, and Jaakko Lehtinen.
\newblock Progressive growing of gans for improved quality, stability, and variation.
\newblock {\em ICLR}, 2018.

\bibitem{karras2019style}
Tero Karras, Samuli Laine, and Timo Aila.
\newblock A style-based generator architecture for generative adversarial networks.
\newblock In {\em CVPR}, 2019.

\bibitem{kim2024instantfamily}
Chanran Kim, Jeongin Lee, Shichang Joung, Bongmo Kim, and Yeul-Min Baek.
\newblock Instantfamily: Masked attention for zero-shot multi-id image generation.
\newblock {\em arXiv preprint arXiv:2404.19427}, 2024.

\bibitem{kim2022adaface}
Minchul Kim, Anil~K Jain, and Xiaoming Liu.
\newblock Adaface: Quality adaptive margin for face recognition.
\newblock In {\em CVPR}, 2022.

\bibitem{flux2024}
Black~Forest Labs.
\newblock Flux.
\newblock \url{https://github.com/black-forest-labs/flux}, 2024.

\bibitem{fluxkrea}
Black~Forest Labs.
\newblock Flux.1 krea.
\newblock \url{https://huggingface.co/black-forest-labs/FLUX.1-Krea-dev}, 2025.

\bibitem{liu2015faceattributes}
Ziwei Liu, Ping Luo, Xiaogang Wang, and Xiaoou Tang.
\newblock Deep learning face attributes in the wild.
\newblock In {\em ICCV}, 2015.

\bibitem{mou2025dreamo}
Chong Mou, Yanze Wu, Wenxu Wu, Zinan Guo, Pengze Zhang, Yufeng Cheng, Yiming Luo, Fei Ding, Shiwen Zhang, Xinghui Li, et~al.
\newblock Dreamo: A unified framework for image customization.
\newblock {\em SIGGRAPH Asia}, 2025.

\bibitem{oord2018representation}
Aaron van~den Oord, Yazhe Li, and Oriol Vinyals.
\newblock Representation learning with contrastive predictive coding.
\newblock {\em arXiv preprint arXiv:1807.03748}, 2018.

\bibitem{openai2025gpt4o}
OpenAI.
\newblock Addendum to gpt-4o system card: Native image generation, 2025.

\bibitem{oquab2023dinov2}
Maxime Oquab, Timothée Darcet, Theo Moutakanni, Huy~V. Vo, Marc Szafraniec, Vasil Khalidov, Pierre Fernandez, Daniel Haziza, Francisco Massa, Alaaeldin El-Nouby, Russell Howes, Po-Yao Huang, Hu~Xu, Vasu Sharma, Shang-Wen Li, Wojciech Galuba, Mike Rabbat, Mido Assran, Nicolas Ballas, Gabriel Synnaeve, Ishan Misra, Herve Jegou, Julien Mairal, Patrick Labatut, Armand Joulin, and Piotr Bojanowski.
\newblock Dinov2: Learning robust visual features without supervision.
\newblock {\em arXiv:2304.07193}, 2023.

\bibitem{pan2023renderme}
Dongwei Pan, Long Zhuo, Jingtan Piao, Huiwen Luo, Wei Cheng, Yuxin Wang, Siming Fan, Shengqi Liu, Lei Yang, Bo~Dai, et~al.
\newblock Renderme-360: A large digital asset library and benchmarks towards high-fidelity head avatars.
\newblock {\em NeurIPS}, 2023.

\bibitem{papantoniou2024arc2face}
Foivos~Paraperas Papantoniou, Alexandros Lattas, Stylianos Moschoglou, Jiankang Deng, Bernhard Kainz, and Stefanos Zafeiriou.
\newblock Arc2face: A foundation model for id-consistent human faces.
\newblock In {\em ECCV}, 2024.

\bibitem{parmar2025object}
Gaurav Parmar, Or~Patashnik, Kuan-Chieh Wang, Daniil Ostashev, Srinivasa Narasimhan, Jun-Yan Zhu, Daniel Cohen-Or, and Kfir Aberman.
\newblock Object-level visual prompts for compositional image generation.
\newblock {\em arXiv preprint arXiv:2501.01424}, 2025.

\bibitem{patashnik2025nested}
Or~Patashnik, Rinon Gal, Daniil Ostashev, Sergey Tulyakov, Kfir Aberman, and Daniel Cohen-Or.
\newblock Nested attention: Semantic-aware attention values for concept personalization.
\newblock In {\em SIGGRAPH}, 2025.

\bibitem{peebles2023scalable}
William Peebles and Saining Xie.
\newblock Scalable diffusion models with transformers.
\newblock In {\em ICCV}, 2023.

\bibitem{peng2024portraitbooth}
Xu~Peng, Junwei Zhu, Boyuan Jiang, Ying Tai, Donghao Luo, Jiangning Zhang, Wei Lin, Taisong Jin, Chengjie Wang, and Rongrong Ji.
\newblock Portraitbooth: A versatile portrait model for fast identity-preserved personalization.
\newblock In {\em CVPR}, 2024.

\bibitem{qian2024omni}
Guocheng Qian, Kuan-Chieh Wang, Or~Patashnik, Negin Heravi, Daniil Ostashev, Sergey Tulyakov, Daniel Cohen-Or, and Kfir Aberman.
\newblock Omni-id: Holistic identity representation designed for generative tasks.
\newblock {\em CVPR}, 2025.

\bibitem{radford2021clip}
Alec Radford, Jong~Wook Kim, Chris Hallacy, Aditya Ramesh, Gabriel Goh, Sandhini Agarwal, Girish Sastry, Amanda Askell, Pamela Mishkin, Jack Clark, et~al.
\newblock Learning transferable visual models from natural language supervision.
\newblock In {\em ICML}, 2021.

\bibitem{ren2023pbidr}
Xingyu Ren, Alexandros Lattas, Baris Gecer, Jiankang Deng, Chao Ma, and Xiaokang Yang.
\newblock Facial geometric detail recovery via implicit representation.
\newblock In {\em FG}, 2023.

\bibitem{ronneberger2015u}
Olaf Ronneberger, Philipp Fischer, and Thomas Brox.
\newblock U-net: Convolutional networks for biomedical image segmentation.
\newblock In {\em MICCAI}, 2015.

\bibitem{ruiz2023dreambooth}
Nataniel Ruiz, Yuanzhen Li, Varun Jampani, Yael Pritch, Michael Rubinstein, and Kfir Aberman.
\newblock Dreambooth: Fine tuning text-to-image diffusion models for subject-driven generation.
\newblock In {\em CVPR}, 2023.

\bibitem{schroff2015facenet}
Florian Schroff, Dmitry Kalenichenko, and James Philbin.
\newblock Facenet: A unified embedding for face recognition and clustering.
\newblock In {\em CVPR}, 2015.

\bibitem{schubert2017dbscan}
Erich Schubert, J{\"o}rg Sander, Martin Ester, Hans~Peter Kriegel, and Xiaowei Xu.
\newblock Dbscan revisited, revisited: why and how you should (still) use dbscan.
\newblock {\em TODS}, 2017.

\bibitem{stacchio2020imago}
Lorenzo Stacchio, Alessia Angeli, Giuseppe Lisanti, Daniela Calanca, and Gustavo Marfia.
\newblock Imago: A family photo album dataset for a socio-historical analysis of the twentieth century.
\newblock {\em arXiv preprint arXiv:2012.01955}, 2020.

\bibitem{valevski2023face0}
Dani Valevski, Danny Lumen, Yossi Matias, and Yaniv Leviathan.
\newblock Face0: Instantaneously conditioning a text-to-image model on a face.
\newblock In {\em SIGGRAPH Asia}, 2023.

\bibitem{wang2024stableidentity}
Qinghe Wang, Xu~Jia, Xiaomin Li, Taiqing Li, Liqian Ma, Yunzhi Zhuge, and Huchuan Lu.
\newblock Stableidentity: Inserting anybody into anywhere at first sight.
\newblock {\em TMM}, 2025.

\bibitem{wang2024instantid}
Qixun Wang, Xu~Bai, Haofan Wang, Zekui Qin, and Anthony Chen.
\newblock Instantid: Zero-shot identity-preserving generation in seconds.
\newblock {\em arXiv preprint arXiv:2401.07519}, 2024.

\bibitem{wang2025faceid}
Shuhe Wang, Xiaoya Li, Jiwei Li, Guoyin Wang, Xiaofei Sun, Bob Zhu, Han Qiu, Mo~Yu, Shengjie Shen, Tianwei Zhang, et~al.
\newblock Faceid-6m: A large-scale, open-source faceid customization dataset.
\newblock {\em arXiv preprint arXiv:2503.07091}, 2025.

\bibitem{wang2024high}
Yibin Wang, Weizhong Zhang, Jianwei Zheng, and Cheng Jin.
\newblock High-fidelity person-centric subject-to-image synthesis.
\newblock In {\em CVPR}, 2024.

\bibitem{wu2025qwenimage}
Chenfei Wu, Jiahao Li, Jingren Zhou, Junyang Lin, Kaiyuan Gao, Kun Yan, Sheng ming Yin, Shuai Bai, Xiao Xu, Yilei Chen, Yuxiang Chen, Zecheng Tang, Zekai Zhang, Zhengyi Wang, An~Yang, Bowen Yu, Chen Cheng, Dayiheng Liu, Deqing Li, Hang Zhang, Hao Meng, Hu~Wei, Jingyuan Ni, Kai Chen, Kuan Cao, Liang Peng, Lin Qu, Minggang Wu, Peng Wang, Shuting Yu, Tingkun Wen, Wensen Feng, Xiaoxiao Xu, Yi~Wang, Yichang Zhang, Yongqiang Zhu, Yujia Wu, Yuxuan Cai, and Zenan Liu.
\newblock Qwen-image technical report.
\newblock {\em arXiv preprint arXiv:2508.02324}, 2025.

\bibitem{wu2025omnigen2}
Chenyuan Wu, Pengfei Zheng, Ruiran Yan, Shitao Xiao, Xin Luo, Yueze Wang, Wanli Li, Xiyan Jiang, Yexin Liu, Junjie Zhou, Ze~Liu, Ziyi Xia, Chaofan Li, Haoge Deng, Jiahao Wang, Kun Luo, Bo~Zhang, Defu Lian, Xinlong Wang, Zhongyuan Wang, Tiejun Huang, and Zheng Liu.
\newblock Omnigen2: Exploration to advanced multimodal generation.
\newblock {\em arXiv preprint arXiv:2506.18871}, 2025.

\bibitem{wu2025uso}
Shaojin Wu, Mengqi Huang, Yufeng Cheng, Wenxu Wu, Jiahe Tian, Yiming Luo, Fei Ding, and Qian He.
\newblock Uso: Unified style and subject-driven generation via disentangled and reward learning.
\newblock {\em arXiv preprint arXiv:2508.18966}, 2025.

\bibitem{wu2025uno}
Shaojin Wu, Mengqi Huang, Wenxu Wu, Yufeng Cheng, Fei Ding, and Qian He.
\newblock Less-to-more generalization: Unlocking more controllability by in-context generation.
\newblock {\em ICCV}, 2025.

\bibitem{wu2024fiva}
Tong Wu, Yinghao Xu, Ryan Po, Mengchen Zhang, Guandao Yang, Jiaqi Wang, Ziwei Liu, Dahua Lin, and Gordon Wetzstein.
\newblock Fiva: Fine-grained visual attribute dataset for text-to-image diffusion models.
\newblock {\em NeurIPS}, 2024.

\bibitem{wu2024infinite}
Yi~Wu, Ziqiang Li, Heliang Zheng, Chaoyue Wang, and Bin Li.
\newblock Infinite-id: Identity-preserved personalization via id-semantics decoupling paradigm.
\newblock In {\em ECCV}, 2024.

\bibitem{xiao2024customsketching}
Chufeng Xiao and Hongbo Fu.
\newblock Customsketching: Sketch concept extraction for sketch-based image synthesis and editing.
\newblock In {\em Computer Graphics Forum}. Wiley Online Library, 2024.

\bibitem{xiao2025fastcomposer}
Guangxuan Xiao, Tianwei Yin, William~T Freeman, Fr{\'e}do Durand, and Song Han.
\newblock Fastcomposer: Tuning-free multi-subject image generation with localized attention.
\newblock {\em IJCV}, 2025.

\bibitem{xiao2024omnigen}
Shitao Xiao, Yueze Wang, Junjie Zhou, Huaying Yuan, Xingrun Xing, Ruiran Yan, Chaofan Li, Shuting Wang, Tiejun Huang, and Zheng Liu.
\newblock Omnigen: Unified image generation.
\newblock {\em arXiv preprint arXiv:2409.11340}, 2024.

\bibitem{xu2024permutation}
Hengyuan Xu, Liyao Xiang, Hangyu Ye, Dixi Yao, Pengzhi Chu, and Baochun Li.
\newblock Permutation equivariance of transformers and its applications.
\newblock In {\em CVPR}, 2024.

\bibitem{yan2023facestudio}
Yuxuan Yan, Chi Zhang, Rui Wang, Yichao Zhou, Gege Zhang, Pei Cheng, Gang Yu, and Bin Fu.
\newblock Facestudio: Put your face everywhere in seconds.
\newblock {\em arXiv preprint arXiv:2312.02663}, 2023.

\bibitem{ye2023ipadapter}
Hu~Ye, Jun Zhang, Sibo Liu, Xiao Han, and Wei Yang.
\newblock Ip-adapter: Text compatible image prompt adapter for text-to-image diffusion models.
\newblock {\em arXiv preprint arxiv:2308.06721}, 2023.

\bibitem{zhai2023siglip}
Xiaohua Zhai, Basil Mustafa, Alexander Kolesnikov, and Lucas Beyer.
\newblock Sigmoid loss for language image pre-training.
\newblock In {\em ICCV}, 2023.

\bibitem{zhang2023adding}
Lvmin Zhang, Anyi Rao, and Maneesh Agrawala.
\newblock Adding conditional control to text-to-image diffusion models.
\newblock In {\em ICCV}, 2023.

\bibitem{zhang2015beyond}
Ning Zhang, Manohar Paluri, Yaniv Taigman, Rob Fergus, and Lubomir Bourdev.
\newblock Beyond frontal faces: Improving person recognition using multiple cues.
\newblock In {\em CVPR}, 2015.

\bibitem{zhang2025idpatch}
Yimeng Zhang, Tiancheng Zhi, Jing Liu, Shen Sang, Liming Jiang, Qing Yan, Sijia Liu, and Linjie Luo.
\newblock Id-patch: Robust id association for group photo personalization.
\newblock In {\em CVPR}, 2025.

\bibitem{zhang2023recognize}
Youcai Zhang, Xinyu Huang, Jinyu Ma, Zhaoyang Li, Zhaochuan Luo, Yanchun Xie, Yuzhuo Qin, Tong Luo, Yaqian Li, Shilong Liu, et~al.
\newblock Recognize anything: A strong image tagging model.
\newblock {\em arXiv preprint arXiv:2306.03514}, 2023.

\bibitem{zhong2018compact}
Yujie Zhong, Relja Arandjelovic, and Andrew Zisserman.
\newblock Compact deep aggregation for set retrieval.
\newblock In {\em ECCV}, 2018.

\bibitem{zhuang2025vistorybench}
Cailin Zhuang, Ailin Huang, Wei Cheng, Jingwei Wu, Yaoqi Hu, Jiaqi Liao, Hongyuan Wang, Xinyao Liao, Weiwei Cai, Hengyuan Xu, et~al.
\newblock Vistorybench: Comprehensive benchmark suite for story visualization.
\newblock {\em arXiv preprint arXiv:2505.24862}, 2025.

\end{thebibliography}
